\pdfoutput=1
\documentclass[10pt,twocolumn,letterpaper]{article}

\usepackage[pagenumbers]{cvpr}

\usepackage{microtype}

\renewcommand{\paragraph}[1]{\vspace{.5em}\noindent\textbf{#1}}

\usepackage{array}
\usepackage{placeins}
\usepackage{tikz}
\usepackage{pgfplots}
\pgfplotsset{compat=1.16}
\usepackage{colortbl}
\usetikzlibrary{positioning,arrows.meta,calc,shadows}
\usepgfplotslibrary{groupplots}
\definecolor{diag}{HTML}{B5172E}%
\definecolor{search}{HTML}{3D4451}%
\definecolor{verify}{HTML}{2E86AB}%

\definecolor{stBox}{HTML}{D1495B}%
\definecolor{stName}{HTML}{E8A33D}%
\definecolor{stPair}{HTML}{6C4AB6}%
\definecolor{stPred}{HTML}{2E86AB}%
\definecolor{stRank}{HTML}{7CC08A}%
\definecolor{panelGrey}{HTML}{F0F0F0}%
\definecolor{edgeGrey}{HTML}{BDBDBD}%
\definecolor{rohared}{HTML}{B5172E}%
\definecolor{rohagreen}{HTML}{2E7D32}
\definecolor{rohalink}{HTML}{9E1B2E}%
\newcommand{\robotagent}[4]{%
\begin{scope}[shift={(#1,#2)}]
  \fill[black, opacity=0.10] (0.03,-0.52) ellipse (0.42 and 0.06);
  \draw[#3!80!black, line width=0.9pt] (0,0.36) -- (0,0.5); \fill[#3!80!black] (0,0.54) circle (0.055);
  \draw[#3!80!black, line width=0.9pt, fill=#3!14, rounded corners=4pt] (-0.42,-0.04) rectangle (0.42,0.38);
  \draw[#3!60!black, line width=0.5pt, fill=white, rounded corners=3pt] (-0.31,0.03) rectangle (0.31,0.3);
  \fill[#3!80!black] (-0.12,0.18) circle (0.05) (0.12,0.18) circle (0.05);
  \draw[#3!80!black, line width=0.6pt] (-0.07,0.09) .. controls (-0.03,0.06) and (0.03,0.06) .. (0.07,0.09);
  \fill[#3!80!black] (-0.47,0.1) rectangle (-0.42,0.24) (0.42,0.1) rectangle (0.47,0.24);
  \draw[#3!80!black, line width=0.9pt, fill=#3!22, rounded corners=4pt] (-0.3,-0.46) rectangle (0.3,-0.08);
  \ifnum#4=1
    \draw[#3!80!black, line width=0.9pt, fill=white] (-0.03,-0.24) circle (0.09); \draw[#3!80!black, line width=1.4pt, line cap=round] (0.04,-0.31) -- (0.12,-0.39);
  \else
    \node[font=\ttfamily\bfseries\tiny, text=#3!80!black, inner sep=0pt] at (0,-0.27) {</>};
  \fi
\end{scope}}
\newcommand{\saybubble}[4]{%
  \node[draw=#3!70!black, fill=white, rounded corners=5pt, line width=0.6pt, font=\sffamily\tiny\itshape, text=black!75, inner sep=2.5pt, align=center, drop shadow={opacity=0.12, shadow xshift=0.5pt, shadow yshift=-0.5pt}] (bub) at (#1,#2) {#4};
  \fill[white] ($(bub.south)+(-0.08,0.02)$) -- ($(bub.south)+(0.08,0.02)$) -- ($(bub.south)+(-0.02,-0.16)$) -- cycle;
  \draw[#3!70!black, line width=0.6pt] ($(bub.south)+(-0.08,0.0)$) -- ($(bub.south)+(-0.02,-0.16)$) -- ($(bub.south)+(0.08,0.0)$);}

\newcommand{\papertitle}{Fix Where It Fails:\texorpdfstring{\\}{ }\texorpdfstring{\underline{Ro}ot-Cause-Guided Agentic \underline{Ha}rness Optimization for \underline{S}cene \underline{G}raph \underline{G}eneration}{Root-Cause-Guided Agentic Harness Optimization for Scene Graph Generation}}
\newcommand{\methodname}{RoHa-SGG\xspace}
\newcommand{\gc}{\textsc{gc}\xspace}
\newcommand{\searchsplit}{search\xspace}
\newcommand{\interimsplit}{selection\xspace}
\newcommand{\testsplit}{test\xspace}
\newlength{\dupw}
\newcommand{\dup}[1]{\makebox[\dupw][l]{\scriptsize\color{rohagreen}(#1)}}
\newcommand{\dnone}{\,\makebox[\dupw]{}}%

\newcommand{\numImagesLossMapRoneSGGPSGSearch}{790\xspace}
\newcommand{\numImagesLossMapRoneSGGVGSearch}{3{,}000\xspace}

\makeatletter
\def\@maketitle{%
   \newpage\null\vskip .3in
   \begin{center}
      {\Large \bf \@title \par}
      \vspace*{18pt}{\large\lineskip .5em
        \begin{tabular}[t]{c}\@author\end{tabular}\par}
      \vskip 6pt
   \end{center}}
\def\abstract{\centerline{\large\bf Abstract}\vspace*{6pt}\noindent\it\ignorespaces}

\makeatother
\usepackage[pagebackref,breaklinks,colorlinks,allcolors=rohalink]{hyperref}

\title{\papertitle}

\author{Jee Eun Kim\qquad Donghyeon Choi\qquad Sangjune An\qquad Hae Beom Lee$^{\dagger}$\qquad Kyong Hwan Jin$^{\dagger}$\\
Korea University\\
{\tt\small \{kimjeeeun98, dong\char`\_choi, ics8088, haebeomlee, kyong\char`\_jin\}@korea.ac.kr}}

\begin{document}
\twocolumn[{%
\renewcommand\twocolumn[1][]{#1}%
\maketitle
\providecommand{\tzLock}[3]{%
\begin{scope}[shift={(#1,#2)}, scale=#3]
  \draw[black!60, line width=0.9pt] (-0.09,0.02) -- (-0.09,0.12) arc (180:0:0.09) -- (0.09,0.02);
  \fill[black!60, rounded corners=0.6pt] (-0.15,-0.16) rectangle (0.15,0.04);
  \fill[white] (0,-0.06) circle (0.035);
\end{scope}}
\providecommand{\tzEval}[3]{%
\begin{scope}[shift={(#1,#2)}, scale=#3]
  \fill[black, opacity=0.10] (0.03,-0.52) ellipse (0.42 and 0.06);
  \draw[verify!80!black, line width=0.9pt, fill=white, rounded corners=3pt] (-0.42,-0.36) rectangle (0.30,0.30);
  \fill[verify!35] (-0.36,-0.30) -- (-0.16,0.0) -- (-0.02,-0.14) -- (0.08,-0.04) -- (0.24,-0.30) -- cycle;
  \fill[stName] (0.10,0.14) circle (0.065);
  \fill[white] (0.30,-0.30) circle (0.22);
  \fill[verify] (0.30,-0.30) circle (0.19);
  \draw[white, line width=1.3pt, line cap=round, line join=round] (0.20,-0.30) -- (0.27,-0.38) -- (0.41,-0.20);
\end{scope}}
\providecommand{\tzStage}[5]{%
\begin{scope}[shift={(#2,#3)}, scale=#4]
  \ifcase#1\or
    \draw[#5, line width=0.8pt] (-0.11,-0.075) rectangle (0.11,0.075);
  \or
    \fill[#5] (-0.12,-0.065) -- (0.045,-0.065) -- (0.12,0) -- (0.045,0.065) -- (-0.12,0.065) -- cycle;\fill[white] (-0.07,0) circle (0.02);
  \or
    \draw[#5, line width=0.8pt] (-0.10,-0.045) -- (0.10,0.045);\fill[#5] (-0.10,-0.045) circle (0.04) (0.10,0.045) circle (0.04);
  \or
    \draw[#5, line width=0.9pt, -{Stealth[length=1.2mm]}] (-0.13,0) -- (0.13,0);
  \or
    \foreach \i/\l in {0/0.22,1/0.16,2/0.10} \fill[#5] (-0.11,0.05-\i*0.06) rectangle ++(\l,0.035);
  \fi
\end{scope}}

{\centering
\resizebox{0.98\textwidth}{!}{%
\begin{tikzpicture}[font=\sffamily\small, >={Stealth[length=2.4mm]},
  ttl/.style={font=\sffamily\bfseries, anchor=north west, text=black!85, inner sep=0pt},
  sub/.style={font=\sffamily\scriptsize, text=black!60, inner sep=0pt, anchor=north west, align=left},
  lab/.style={font=\sffamily\scriptsize, align=center, text=black!70, inner sep=1pt},
  nm/.style={font=\sffamily\scriptsize\bfseries, inner sep=0pt, anchor=north},
  chip/.style={minimum width=0.38cm, minimum height=0.38cm, inner sep=0pt, rounded corners=2pt},
  card/.style={draw=#1!70!black, fill=#1, line width=0.9pt, rounded corners=2.5pt, minimum width=1.05cm, minimum height=0.44cm, inner sep=0pt, drop shadow={opacity=0.16, shadow xshift=0.6pt, shadow yshift=-0.6pt}},
  fix/.style={draw=#1!70!black, fill=#1, line width=0.8pt, rounded corners=2pt, minimum width=0.44cm, minimum height=0.36cm, inner sep=0pt},
  bdg/.style={circle, fill=#1!80!black, draw=white, line width=0.6pt, text=white, inner sep=0pt, minimum size=0.23cm, font=\sffamily\bfseries\fontsize{5.4}{5.4}\selectfont},
  tiny/.style={font=\sffamily\bfseries\fontsize{5.8}{6}\selectfont, text=white, inner sep=0pt, anchor=west},
]
\fill[rounded corners=7pt, black!3] (0,-1.75) rectangle (3.75,2.5);
\fill[rounded corners=7pt, stPair!6!white] (4.0,-1.75) rectangle (11.0,2.5);
\fill[rounded corners=7pt, rohared!5!white] (11.25,-1.75) rectangle (18.2,2.5);
\node[ttl] at (0.2,2.3) {Problem};
\node[sub] at (0.2,1.87) {where each model fails};
\node[ttl, text=search!90!black] at (4.2,2.3) {Harness search {\mdseries\small\color{black!70}(offline)}};
\node[ttl, text=rohared] at (11.45,2.3) {Result {\mdseries\small\color{black!70}(deployed)}};
\begin{scope}[yshift=-0.45cm]
\foreach \x/\w/\c in {1.55/box/stBox,2.0/name/stName,2.45/pair/stPair,2.9/pred./stPred,3.35/rank/stRank} \node[font=\sffamily\bfseries\fontsize{4.7}{5}\selectfont, text=\c!85!black, inner sep=0pt, anchor=base] at (\x,1.41) {\w};
\foreach \y/\n/\f in {0.95/{two-stage\\model}/3, 0.15/{one-stage\\model}/4, -0.65/{MLLM}/1} {
  \node[lab, align=right, anchor=east, text=black!85] at (1.3,\y) {\n};
  \draw[black!30, line width=1pt] (1.4,\y) -- (3.5,\y);
  \foreach \c [count=\k from 1] in {stBox,stName,stPair,stPred,stRank} {
    \pgfmathsetmacro{\xx}{1.1+\k*0.45}
    \ifnum\k=\f
      \node[chip, fill=\c, draw=\c!70!black, line width=0.7pt] at (\xx,\y) {};
    \else\ifnum\k=5 \ifnum\f>1
      \node[chip, fill=stRank!85!black, draw=stRank!60!black, line width=0.7pt] at (\xx,\y) {};
      \else
      \node[chip, fill=black!8, draw=black!22, line width=0.5pt] at (\xx,\y) {};
      \fi
    \else
      \node[chip, fill=black!8, draw=black!22, line width=0.5pt] at (\xx,\y) {};
    \fi\fi
  }
}
\end{scope}
\begin{scope}[shift={(4.85,0.35)}, scale=0.85]\robotagent{0}{0}{diag}{1}\end{scope}
\saybubble{4.85}{1.42}{diag}{``where does\\it fail?''}
\node[nm, text=diag!90!black] at (4.85,-0.27) {RCA agent};
\foreach \y/\k/\c/\s/\n in {1.15/3/stPair/pair/1, 0.57/4/stPred/pred./2, -0.01/1/stBox/box/3} {
  \node[card=\c] (k\n) at (6.65,\y) {};
  \tzStage{\k}{6.37}{\y}{1.35}{white}
  \node[tiny=\c] at (6.6,\y) {\s};
  \node[bdg=\c] at (k\n.north west) {\n};
}
\colorlet{stRankD}{stRank!80!black}
\node[card=stRankD] (k4) at (6.65,-0.59) {};
\node[lab, text=black!65] at (6.65,-0.98) {causes};
\tzStage{5}{6.37}{-0.59}{1.35}{white}
\node[tiny=stRankD] at (6.6,-0.59) {rank};
\node[bdg=stRankD] at (k4.north west) {4};
\begin{scope}[shift={(8.5,0.35)}, scale=0.85]\robotagent{0}{0}{search}{2}\end{scope}
\saybubble{8.5}{1.42}{search}{``fix each\\cause''}
\node[nm, text=search!90!black] at (8.5,-0.27) {proposer agent};
\fill[white, draw=verify!45, line width=0.7pt, rounded corners=1.5pt] (10.13,0.03) rectangle (10.63,0.71);
\fill[verify!8, draw=verify!85!black, line width=0.9pt, rounded corners=1.5pt] (9.97,-0.07) rectangle (10.47,0.61);
\node[font=\ttfamily\bfseries\fontsize{8}{8}\selectfont, text=verify!85!black] at (10.22,0.27) {</>};
\node[nm, text=verify!80!black] at (10.25,-0.27) {harness};
\draw[->, line width=1.8pt, diag!85!black] (5.3,0.35) -- (5.95,0.35);
\draw[->, line width=1.8pt, diag!85!black] (7.33,0.35) -- (7.95,0.35);
\draw[->, line width=1.3pt, black!50] (9.05,0.35) -- (9.75,0.35);
\node[font=\sffamily\bfseries\fontsize{6}{7}\selectfont, text=black!65, anchor=south, inner sep=1.5pt] at (9.4,0.38) {fix};
\draw[->, line width=1.5pt, black!55, rounded corners=10pt] (10.25,-0.55) -- (10.25,-1.32) -- (4.85,-1.32) -- (4.85,-0.58);
\node[fill=stPair!6!white, inner sep=3pt, font=\sffamily\bfseries\scriptsize, text=black!70] at (7.55,-1.32) {evaluate \& re-diagnose};
\draw[->, line width=2.6pt, rohared] (11.0,0.35) -- (11.415,0.35);%
\begin{scope}[yshift=-0.25cm]
\def\XO{13.5}\def\W{0.46}
\pgfmathsetmacro{\Xb}{\XO+0.5*\W}\pgfmathsetmacro{\Xn}{\XO+1.5*\W}\pgfmathsetmacro{\Xp}{\XO+2.5*\W}\pgfmathsetmacro{\Xq}{\XO+3.5*\W}\pgfmathsetmacro{\Xr}{\XO+4.5*\W}
\draw[rohared, line width=1.2pt, dash pattern=on 3pt off 1.6pt, fill=white, rounded corners=5pt] (11.45,-1.05) rectangle (15.9,1.9);
\node[fill=white, inner sep=1.5pt, font=\sffamily\bfseries\scriptsize, text=rohared, anchor=west] at (11.6,1.9) {selected harness};
\foreach \x/\k/\c in {\Xb/1/stBox,\Xn/2/stName,\Xp/3/stPair,\Xq/4/stPred,\Xr/5/stRank} {\tzStage{\k}{\x}{1.48}{1.1}{\c!85!black}}
\foreach \y/\f/\m in {0.95/two-stage model/REACT++, 0.15/one-stage model/EGTR, -0.65/MLLM/R1-SGG} {
  \tzLock{11.68}{\y+0.02}{0.75}
  \node[font=\sffamily\bfseries\fontsize{5.8}{6.6}\selectfont, text=black!85, anchor=west, inner sep=0pt] at (11.82,\y+0.09) {\f};
  \node[font=\sffamily\fontsize{5.2}{6}\selectfont, text=black!55, anchor=west, inner sep=0pt] at (11.82,\y-0.14) {\m};
  \fill[panelGrey, rounded corners=3pt] (\XO+0.03,\y-0.27) rectangle (\XO+5*\W-0.03,\y+0.27);
  \foreach \x/\c in {\Xb/stBox,\Xn/stName,\Xp/stPair,\Xq/stPred,\Xr/stRank} \fill[\c!60] (\x,\y) circle (0.045);
}
\foreach \x/\y/\k/\c/\n in {\Xp/0.95/3/stPair/1, \Xq/0.15/4/stPred/2, \Xb/-0.65/1/stBox/3} {
  \node[fix=\c] (f\n) at (\x,\y) {};
  \tzStage{\k}{\x}{\y}{1.1}{white}
  \node[bdg=\c] at (f\n.north west) {\n};
}
\foreach \y/\bc/\bn in {0.95/stRankD/4, 0.15/stRankD/4, -0.65/stRankD/4} {\node[fix=stRankD] (fr) at (\Xr,\y) {}; \tzStage{5}{\Xr}{\y}{1.1}{white} \node[bdg=\bc] at (fr.north west) {\bn};}
\def\YA{-0.72}\def\SC{0.048}
\foreach \x/\a/\b/\l in {16.55/24.1/38.7/R@100, 17.55/6.3/15.2/mR@100} {
  \fill[black!35] (\x-0.4,\YA) rectangle (\x-0.03,\YA+\a*\SC);
  \fill[rohared] (\x+0.03,\YA) rectangle (\x+0.4,\YA+\b*\SC);
  \node[font=\sffamily\fontsize{5.2}{6}\selectfont, text=black!70, anchor=south, inner sep=1pt] at (\x-0.215,\YA+\a*\SC) {\a};
  \node[font=\sffamily\bfseries\fontsize{5.4}{6}\selectfont, text=rohared, anchor=south, inner sep=1pt] at (\x+0.215,\YA+\b*\SC) {\b};
  \node[font=\sffamily\bfseries\fontsize{6}{7}\selectfont, text=black!80, anchor=north, inner sep=0pt] at (\x,\YA-0.08) {\l};
}
\draw[black!50, line width=0.6pt] (16.08,\YA) -- (17.97,\YA);
\node[font=\sffamily\fontsize{5.8}{7}\selectfont, text=black!65, inner sep=0pt] at (17.05,-1.18) {R1-SGG, VG150 test};
\fill[black!35] (15.98,1.62) rectangle ++(0.15,0.12); \node[font=\sffamily\fontsize{5.2}{6}\selectfont, text=black!70, anchor=west, inner sep=0pt] at (16.17,1.68) {baseline};
\fill[rohared] (17.0,1.62) rectangle ++(0.15,0.12); \node[font=\sffamily\fontsize{5.2}{6}\selectfont, text=black!70, anchor=west, inner sep=0pt] at (17.19,1.68) {RoHa-SGG};
\end{scope}
\end{tikzpicture}}

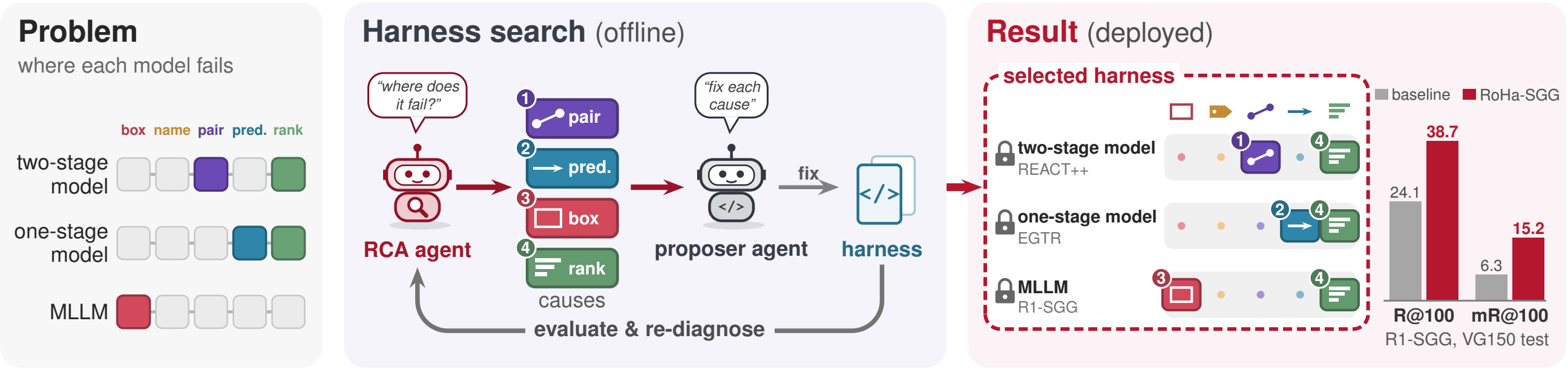
\captionof{figure}{\textbf{Frozen SGG models fail at different components. \methodname{} fixes each model without retraining.} A root-cause analysis (RCA) agent finds the components at which each model fails (left). A proposer agent turns each cause into a candidate harness, which is evaluated before the analysis is repeated. The selected harness edits these components and raises \gc R@100 and mR@100. The bars show R1-SGG with the multi-pass harness on the VG150 \testsplit set.}
\label{fig:teaser}
\par}
\vspace{10pt}

}]
{\renewcommand{\thefootnote}{}\footnotetext{$^{\dagger}$Corresponding authors.}}
\begin{abstract}
Scene graph generation (SGG) aims to detect objects in an image and predict the relationships between them.
The standard PredCls, SGCls, and SGDet settings show how much performance drops without ground-truth boxes and labels, but not where or why a model fails.
Knowing this would allow fixing a model where it fails, without retraining.
We propose \textbf{\methodname} (\textbf{\underline{Ro}}ot-cause-guided \textbf{\underline{Ha}}rness optimization for \textbf{\underline{SGG}}), which extends agentic harness optimization to SGG. It optimizes the harness, the inference code around a frozen SGG model.
\methodname{} has two main LLM agents. A root-cause analysis (RCA) agent finds the components where the model fails and diagnoses why, and a proposer agent searches for a harness that fixes these causes.
We evaluate \methodname{} on Visual Genome and PSG with representative two-stage, one-stage, and multimodal-LLM SGG models.
The analysis shows that the components where ground-truth triplets are lost differ across models and datasets.
\methodname{} improves every metric on all five model--dataset combinations. Graph-constrained R@100 and mR@100 rise by up to 15.8 and 26.4 points for the multimodal LLM and by up to 3.6 and 10.1 points for the two-stage and one-stage models.
With root-cause guidance, the harness search reaches the final score of an unguided search up to three times faster.
\end{abstract}

\section{Introduction}
\label{sec:intro}

Scene graph generation (SGG) describes an image as a set of (subject, predicate, object) triplets, such as (man, riding, horse), over the objects it contains~\cite{xu2017imp,krishna2017vg}.
Most SGG models follow one of three architectures: two-stage models first detect objects and then classify a predicate for each candidate pair (REACT++~\cite{neau2026reactpp}), one-stage models predict objects and predicates together in a single DETR-style pass~\cite{carion2020end} (EGTR~\cite{im2024egtr}), and multimodal large language models (MLLMs) write the whole graph as text (R1-SGG~\cite{chen2025r1sgg}).
Whatever the architecture, a ground-truth triplet counts as a match only if it passes five components: a correct box and name for both objects, a proposal for the pair, the correct predicate, and a rank within the top $K$.
SGG is usually evaluated in three settings~\cite{xu2017imp}, which give the model ground-truth boxes and labels (PredCls), boxes only (SGCls), or neither (SGDet). Comparing the three shows how much a model depends on the given boxes and labels, but not at which component it loses ground-truth triplets.

A model can fail at any of the five components, and where it fails differs by model and dataset (\cref{fig:teaser}, left).
Following TIDE~\cite{bolya2020tide} and Diag-HOI~\cite{zhu2023diaghoi}, we measure this with the same procedure for all three models. We correct components with their ground truth (the true boxes, names, pairs and predicates, and an ideal ranking) and attribute the recovered recall to each component (\cref{fig:lossmap}).
In graph-constrained (\gc) R@20, the missed ground-truth triplets spread over several components, and REACT++, EGTR and R1-SGG all lose the most at the pair component. On VG150, the MLLM R1-SGG also loses 27.9 points at the box component, against 7.5 for each classic model. R1-SGG loses nothing at the rank component, where the classic models lose 11.6 to 16.5. On PSG, REACT++ and R1-SGG both lose more at the predicate component than on VG150 (19.6 and 16.6 vs.\ 16.1 and 9.7 points).
A method that targets a single component therefore recovers only part of the missed triplets, and one that helps one model or dataset may not help another.
Earlier SGG analyses measure only parts of this, such as pair recall~\cite{wang2023pairnet}, what the detector can reach~\cite{qu2026reachability} or separate object, pair and predicate scores~\cite{lorenz2024metrics}, or substitute ground truth in a fixed order~\cite{han2021benchmark}. The weaknesses these analyses reveal are then usually fixed by retraining the model, for example with a new pair proposal module~\cite{wang2023pairnet}.

Retraining, however, is costly and must be repeated for every model and dataset.
We instead optimize the \emph{harness}, the inference code that prepares the model's input, calls the model, runs external tools, and post-processes and ranks the output. Because it handles every step from the image to the final scene graph, the harness can reach every component without changing the model.

Harness optimization has recently been shown to work for LLM systems~\cite{lee2026metaharness} and, in concurrent work, for video understanding and embodied agents~\cite{lin2026videoharness,lee2026emho}. It has not yet been applied to SGG.
The space of possible harnesses is large, and scoring each candidate harness means running the model on a large image set. The search is therefore expensive and can afford only a few candidates. These few candidates should target where the model actually fails.
Recent harness optimizers diagnose failures by reading an LLM agent's execution traces, such as its prompts, tool calls and intermediate reasoning~\cite{chen2026harnessfix,park2026autosaddler,wang2026t2a}. A frozen SGG model produces no such trace. Its output shows which ground-truth triplets are missing, but not where or why they were lost. We therefore measure where, component by component, and then test why.

We introduce \textbf{\methodname} (\textbf{\underline{Ro}}ot-cause-guided \textbf{\underline{Ha}}rness optimization for \textbf{\underline{SGG}}), which uses this per-component diagnosis to guide the harness search (\cref{fig:teaser,fig:method}).
It has two LLM agents. A root-cause analysis (RCA) agent finds where and why the model loses ground-truth triplets, using the same procedure for every model and dataset. It splits the missed triplets over the five components with Shapley values and tests each suspected cause with an intervention whose effect it predicts before running it.
A proposer agent then searches for harnesses that fix the diagnosed causes first, rather than exploring the code space without direction.
The search runs in rounds, and before each round the RCA agent analyzes the current best harness again, so the proposer always works on the causes that remain.

We evaluate \methodname{} with REACT++, EGTR and R1-SGG on VG150~\cite{krishna2017vg,xu2017imp} and PSG~\cite{yang2022psg}.
Without retraining the SGG models, the harness found by each search improves every metric on the official test sets of all five model--dataset combinations (\cref{tab:main}).
\gc R@100 and mR@100 rise by up to 15.8 and 26.4 points for R1-SGG and by up to 3.6 and 10.1 points for REACT++ and EGTR.

Diagnosing first also makes the search faster. To measure this, we run three searches with the same loop, adapted from Meta-Harness~\cite{lee2026metaharness}, and the same number of iterations. The unguided search uses only the loop, the knowledge-guided search adds a human-written guide based on the SGG literature, and the RCA-guided search adds the root-cause analysis. On REACT++ and EGTR, the RCA-guided search reaches the final score of the other two searches after only 5 and 4 of 12 iterations (\cref{fig:levels}). Candidate harnesses guided by the diagnosis also reduce the error they target more often. In the first round, 60\% of them lower the error of their target component, compared with 41\% when the targets are reassigned at random.

In summary, our main contributions are as follows:
\begin{itemize}[leftmargin=*,itemsep=1pt,topsep=2pt]
\item We propose \methodname, the first agentic harness optimization for SGG. It improves a frozen SGG model without retraining it by searching over the inference code around it, guided by an analysis of where and why the model fails.
\item We introduce a root-cause analysis for SGG that splits each model's missed ground-truth triplets over five components with order-independent Shapley values and tests suspected causes with interventions. It reveals that models and datasets lose ground-truth triplets at different components.
\item With two-stage, one-stage and multimodal-LLM SGG models on VG150 and PSG, \methodname improves every metric on all five model--dataset combinations. On VG150, root-cause analysis makes the search up to three times faster than unguided and knowledge-guided searches.
\end{itemize}

\providecommand{\tzLock}[3]{%
\begin{scope}[shift={(#1,#2)}, scale=#3]
  \draw[black!60, line width=0.9pt] (-0.09,0.02) -- (-0.09,0.12) arc (180:0:0.09) -- (0.09,0.02);
  \fill[black!60, rounded corners=0.6pt] (-0.15,-0.16) rectangle (0.15,0.04);
  \fill[white] (0,-0.06) circle (0.035);
\end{scope}}
\providecommand{\tzEval}[3]{%
\begin{scope}[shift={(#1,#2)}, scale=#3]
  \fill[black, opacity=0.10] (0.03,-0.52) ellipse (0.42 and 0.06);
  \draw[verify!80!black, line width=0.9pt, fill=white, rounded corners=3pt] (-0.42,-0.36) rectangle (0.30,0.30);
  \fill[verify!35] (-0.36,-0.30) -- (-0.16,0.0) -- (-0.02,-0.14) -- (0.08,-0.04) -- (0.24,-0.30) -- cycle;
  \fill[stName] (0.10,0.14) circle (0.065);
  \fill[white] (0.30,-0.30) circle (0.22);
  \fill[verify] (0.30,-0.30) circle (0.19);
  \draw[white, line width=1.3pt, line cap=round, line join=round] (0.20,-0.30) -- (0.27,-0.38) -- (0.41,-0.20);
\end{scope}}
\providecommand{\tzStage}[5]{%
\begin{scope}[shift={(#2,#3)}, scale=#4]
  \ifcase#1\or
    \draw[#5, line width=0.8pt] (-0.11,-0.075) rectangle (0.11,0.075);
  \or
    \fill[#5] (-0.12,-0.065) -- (0.045,-0.065) -- (0.12,0) -- (0.045,0.065) -- (-0.12,0.065) -- cycle;\fill[white] (-0.07,0) circle (0.02);
  \or
    \draw[#5, line width=0.8pt] (-0.10,-0.045) -- (0.10,0.045);\fill[#5] (-0.10,-0.045) circle (0.04) (0.10,0.045) circle (0.04);
  \or
    \draw[#5, line width=0.9pt, -{Stealth[length=1.2mm]}] (-0.13,0) -- (0.13,0);
  \or
    \foreach \i/\l in {0/0.22,1/0.16,2/0.10} \fill[#5] (-0.11,0.05-\i*0.06) rectangle ++(\l,0.035);
  \fi
\end{scope}}

\tikzset{
  fmttl/.style={font=\sffamily\bfseries, anchor=north west, text=black!85, inner sep=0pt},
  fmnm/.style={font=\sffamily\scriptsize\bfseries, inner sep=0pt, anchor=north},
  fmcard/.style={draw=#1!70!black, fill=#1, line width=0.9pt, rounded corners=2.5pt, minimum width=1.05cm, minimum height=0.44cm, inner sep=0pt, drop shadow={opacity=0.16, shadow xshift=0.6pt, shadow yshift=-0.6pt}},
  fmbdg/.style={circle, fill=#1!80!black, draw=white, line width=0.6pt, text=white, inner sep=0pt, minimum size=0.25cm, font=\sffamily\bfseries\fontsize{5.4}{5.4}\selectfont},
  fmtiny/.style={font=\sffamily\bfseries\fontsize{5.8}{6}\selectfont, text=white, inner sep=0pt, anchor=west},
}
\newcommand{\fmCard}[7]{%
  \node[fmcard=#5] (#1) at (#2,#3) {};
  \tzStage{#4}{#2-0.28}{#3}{1.35}{white}
  \node[fmtiny] at (#2-0.05,#3) {#6};
  \node[fmbdg=#5] at (#1.north west) {#7};}
\newcommand{\fmCards}[3]{%
  \fmCard{cc1}{#1}{#2}{3}{stPair}{pair}{1}
  \fmCard{cc2}{#1}{#2-#3}{4}{stPred}{pred.}{2}
  \fmCard{cc3}{#1}{#2-2*#3}{1}{stBox}{box}{3}}
\newcommand{\fmLossMap}[3]{%
  \foreach \k/\h/\c [count=\i from 0] in {1/0.40/stBox, 2/0.22/stName, 3/0.62/stPair, 4/0.34/stPred, 5/0.12/stRank} {
    \fill[\c, rounded corners=0.8pt] (#1+\i*#3,#2) rectangle ++(0.7*#3,\h);
    \tzStage{\k}{#1+\i*#3+0.35*#3}{#2-0.2}{0.95}{\c!85!black}
  }
  \draw[black!50, line width=0.5pt] (#1-0.1,#2) -- (#1+4.7*#3+0.1,#2);}
\newcommand{\fmImage}[3]{\begin{scope}[shift={(#1,#2)}, scale=#3]
  \draw[black!60, line width=0.8pt, fill=white, rounded corners=1.5pt] (-0.32,-0.25) rectangle (0.32,0.25);
  \fill[black!30] (-0.27,-0.2) -- (-0.08,0.06) -- (0.04,-0.06) -- (0.12,0.02) -- (0.27,-0.2) -- cycle;
  \fill[stName] (0.15,0.12) circle (0.055);
\end{scope}}
\newcommand{\fmGraph}[3]{\begin{scope}[shift={(#1,#2)}, scale=#3]
  \draw[black!60, line width=0.7pt, -{Stealth[length=1.2mm]}] (-0.22,0.12) -- (0.17,0.15);
  \draw[black!60, line width=0.7pt, -{Stealth[length=1.2mm]}] (-0.24,0.06) -- (-0.04,-0.15);
  \draw[black!60, line width=0.7pt, -{Stealth[length=1.2mm]}] (0.24,0.09) -- (0.06,-0.14);
  \fill[stBox] (-0.27,0.12) circle (0.075); \fill[stPair] (0.25,0.15) circle (0.075); \fill[stPred] (0,-0.2) circle (0.075);
\end{scope}}

\newcommand{\robotface}{%
  \draw[white, line width=0.9pt, line cap=round] (0,0.21) -- (0,0.32); \fill[white] (0,0.36) circle (0.06);
  \draw[white, line width=0.9pt, rounded corners=1.8pt] (-0.4,-0.21) rectangle (0.4,0.21);
  \fill[white] (-0.15,0.0) circle (0.075) (0.15,0.0) circle (0.075);}
\newcommand{\fmHarness}[4]{%
  \begin{scope}[shift={(#2,#3)}, scale=#4]
  \node[draw=rohared, line width=1.1pt, dash pattern=on 3pt off 1.6pt, fill=white, rounded corners=4pt, minimum width=0.95cm, minimum height=0.9cm, inner sep=0pt] (#1) at (0,0) {};
  \node[font=\ttfamily\bfseries\fontsize{7}{7}\selectfont, text=rohared, inner sep=0pt] at (0,0.17) {</>};
  \foreach \i/\c in {0/stBox,1/stName,2/stPair,3/stPred,4/stRank} \fill[\c] (-0.355+\i*0.15,-0.28) rectangle ++(0.11,0.11);
  \end{scope}}
\newcommand{\fmBest}[3]{\fmHarness{#1}{#2}{#3}{1}}
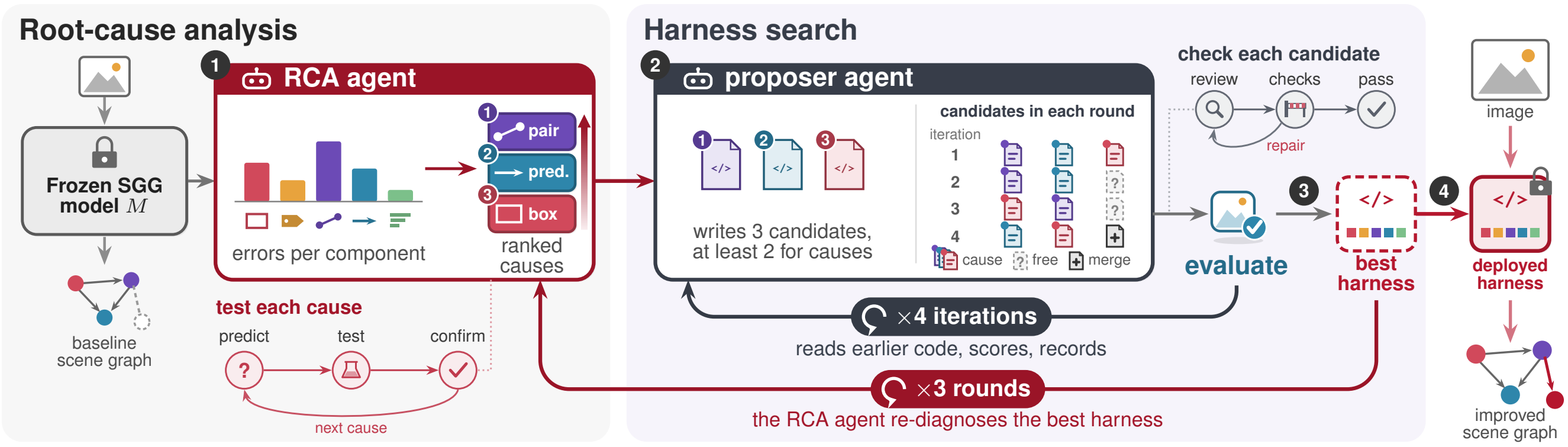
\begin{figure*}[t]
\centering
\resizebox{\textwidth}{!}{%
\begin{tikzpicture}[font=\sffamily\small, >={Stealth[length=2.3mm]},
  abox/.style={draw=#1, line width=1.1pt, rounded corners=6pt, fill=white, drop shadow={opacity=0.13, shadow xshift=0.7pt, shadow yshift=-0.7pt}},
  num/.style={circle, fill=black!80, text=white, font=\sffamily\bfseries\fontsize{6.5}{6.5}\selectfont, inner sep=0pt, minimum size=3.6mm},
  ln/.style={font=\sffamily\scriptsize, inner sep=0pt, text=black!70, align=center},
  flow/.style={->, line width=1.0pt, black!55},
]
\fill[rounded corners=7pt, black!3] (0,0.78) rectangle (7.4,6.1);
\fill[rounded corners=7pt, stPair!6!white] (7.6,0.78) rectangle (17.32,6.1);
\node[fmttl] at (0.2,5.93) {Root-cause analysis};
\node[fmttl, text=search!90!black] at (7.8,5.93) {Harness search};
\node[draw=black!65, fill=black!8, line width=1.0pt, rounded corners=5pt, minimum width=2.0cm, minimum height=1.3cm, drop shadow={opacity=0.15, shadow xshift=0.7pt, shadow yshift=-0.7pt}] (fm) at (1.25,3.95) {};
\fmImage{1.25}{5.22}{0.95}
\draw[->, line width=0.9pt, black!45] (1.25,4.97) -- (1.25,4.63);
\draw[->, line width=0.9pt, black!45] (1.25,3.27) -- (1.25,2.93);
\fmGraph{1.25}{2.55}{1.3}
\begin{scope}[shift={(1.25,2.55)}, scale=1.3]%
  \draw[black!40, line width=0.8pt, dash pattern=on 1.5pt off 1.2pt] (0.27,0.1) -- (0.33,-0.17); \draw[black!40, line width=0.7pt, dash pattern=on 1.2pt off 0.9pt, fill=white] (0.35,-0.24) circle (0.07);
\end{scope}
\node[ln, text=black!70, font=\sffamily\fontsize{6}{7}\selectfont, align=center] at (1.25,1.86) {baseline\\[-1pt]scene graph};
\tzLock{1.25}{4.27}{1.0}
\node[font=\sffamily\bfseries\scriptsize, text=black!88, align=center, inner sep=0pt] at (1.25,3.78) {Frozen SGG\\[-1pt]model $M$};
\draw[flow] (fm.east) -- (2.6,3.95);
\draw[abox=diag!85!black, line width=1.3pt, rounded corners=6pt] (2.6,2.75) rectangle (7.2,5.36);
\fill[diag!85!black, rounded corners=6pt] (2.6,5.36) rectangle (7.2,5.0); \fill[diag!85!black] (2.6,5.0) rectangle (7.2,5.12);
\begin{scope}[shift={(3.1,5.165)}, scale=0.4]\robotface\end{scope}
\node[font=\sffamily\bfseries\small, text=white, anchor=west, inner sep=0pt] at (3.42,5.18) {RCA agent};
\node[num] at (2.6,5.36) {1};
\begin{scope}[shift={(2.95,3.7)}, scale=1.18]\fmLossMap{0}{0}{0.37}\end{scope}
\node[ln] at (3.98,3.05) {errors per component};
\draw[->, line width=1.0pt, diag!85!black] (5.15,4.1) -- (5.78,4.1);
\fmCards{6.45}{4.55}{0.5}
\node[ln] at (6.45,3.05) {ranked\\[-1pt]causes};
\shade[top color=diag!85!black, bottom color=diag!8] (7.05,4.60) rectangle (7.12,3.36);%
\fill[diag!85!black] (7.005,4.60) -- (7.165,4.60) -- (7.085,4.76) -- cycle;
\tikzset{sp/.style={circle, draw=#1!80, fill=#1!8, line width=0.6pt, minimum size=0.45cm, inner sep=0pt}, spl/.style={ln, text=black!78, font=\sffamily\fontsize{5.8}{6.4}\selectfont}, spt/.style={ln, anchor=west, font=\sffamily\bfseries\fontsize{6.8}{7.4}\selectfont}}
\node[spt, text=diag!90!black] at (2.6,2.42) {test each cause};
\draw[diag!45, line width=0.6pt, densely dotted] (5.95,2.75) -- (5.95,1.65) -- (5.775,1.65);
\foreach \n/\x/\t in {cp/2.95/predict, ct/4.25/test, cc/5.55/confirm} {
  \node[sp=diag] (\n) at (\x,1.65) {};
  \node[spl, anchor=base] at (\x,2.0) {\t};
}
\node[font=\sffamily\bfseries\scriptsize, text=diag!85, inner sep=0pt] at (cp) {?};
\begin{scope}[shift={(4.25,1.65)}, scale=0.72]%
  \draw[diag!85, line width=0.6pt, line join=round] (-0.08,0.14) -- (-0.08,0.03) -- (-0.15,-0.12) -- (0.15,-0.12) -- (0.08,0.03) -- (0.08,0.14);
  \fill[diag!45] (-0.12,-0.105) -- (0.12,-0.105) -- (0.07,-0.02) -- (-0.07,-0.02) -- cycle;
  \draw[diag!85, line width=0.6pt] (-0.11,0.14) -- (0.11,0.14);
\end{scope}
\begin{scope}[shift={(5.55,1.65)}, scale=0.72]\draw[diag!85, line width=1.0pt, line cap=round, line join=round] (-0.13,0) -- (-0.04,-0.1) -- (0.14,0.11);\end{scope}
\draw[->, line width=0.6pt, diag!85, >={Stealth[length=1.5mm]}] (cp.east) -- (ct.west);
\draw[->, line width=0.6pt, diag!85, >={Stealth[length=1.5mm]}] (ct.east) -- (cc.west);
\draw[->, line width=0.6pt, diag!75, >={Stealth[length=1.4mm]}] (cc.south) .. controls (5.45,1.0) and (3.05,1.0) .. (cp.south);
\node[spl, text=diag!85, font=\sffamily\fontsize{5.2}{6}\selectfont] at (4.25,0.95) {next cause};
\newcommand{\pgDoc}[5]{%
  \begin{scope}[shift={(#1,#2)}, scale=#3]
  \if#4c\draw[#5!75!black, line width=0.6pt, fill=#5!16] (-0.2,-0.26) -- (-0.2,0.26) -- (0.1,0.26) -- (0.2,0.16) -- (0.2,-0.26) -- cycle;
    \draw[#5!75!black, line width=0.45pt] (0.1,0.26) -- (0.1,0.16) -- (0.2,0.16);
    \draw[#5!80!black, line width=0.7pt, line cap=round] (-0.1,0.02) -- (0.1,0.02) (-0.1,-0.1) -- (0.04,-0.1); \fill[#5] (-0.2,0.26) circle (0.1);\fi
  \if#4f\draw[black!40, line width=0.55pt, fill=black!4, dash pattern=on 1.2pt off 0.9pt] (-0.2,-0.26) -- (-0.2,0.26) -- (0.1,0.26) -- (0.2,0.16) -- (0.2,-0.26) -- cycle;
    \node[font=\sffamily\bfseries\fontsize{4.8}{5}\selectfont, text=black!50, inner sep=0pt] at (0,-0.03) {?};\fi
  \if#4m\draw[black!80, line width=0.6pt, fill=black!10] (-0.2,-0.26) -- (-0.2,0.26) -- (0.1,0.26) -- (0.2,0.16) -- (0.2,-0.26) -- cycle;
    \draw[black!80, line width=0.45pt] (0.1,0.26) -- (0.1,0.16) -- (0.2,0.16);
    \draw[black!80, line width=0.85pt, line cap=round] (-0.09,-0.04) -- (0.09,-0.04) (0,-0.13) -- (0,0.05);\fi
  \end{scope}}
\newcommand{\fmProgT}[4]{%
  \begin{scope}[shift={(#1,#2)}, scale=#3]
  \draw[#4!70!black, line width=0.7pt, fill=#4!14] (-0.2,-0.26) -- (-0.2,0.26) -- (0.1,0.26) -- (0.2,0.16) -- (0.2,-0.26) -- cycle;
  \draw[#4!70!black, line width=0.5pt] (0.1,0.26) -- (0.1,0.16) -- (0.2,0.16);
  \node[font=\ttfamily\bfseries\fontsize{4}{4}\selectfont, text=#4!75!black, inner sep=0pt] at (0,-0.04) {</>};
  \end{scope}}
\draw[->, line width=1.3pt, diag!85!black] (7.2,3.95) -- (7.95,3.95);
\draw[abox=search!90!black, line width=1.3pt, rounded corners=6pt] (7.95,2.75) rectangle (14.0,5.36);
\fill[search!90!black, rounded corners=6pt] (7.95,5.36) rectangle (14.0,5.0); \fill[search!90!black] (7.95,5.0) rectangle (14.0,5.12);
\begin{scope}[shift={(8.47,5.165)}, scale=0.4]\robotface\end{scope}
\node[font=\sffamily\bfseries\small, text=white, anchor=west, inner sep=0pt] at (8.79,5.18) {proposer agent};
\node[num] at (7.95,5.36) {2};
\foreach \x/\c/\b in {8.75/stPair/1, 9.5/stPred/2, 10.25/stBox/3} {
  \fmProgT{\x}{4.15}{1.15}{\c}
  \node[fmbdg=\c] at (\x-0.23,4.45) {\b};
}
\node[ln] at (9.5,3.22) {writes 3 candidates,\\[-1pt]at least 2 for causes};
\draw[black!14, line width=0.5pt] (11.12,2.95) -- (11.12,4.88);
\node[ln, text=search!90!black, font=\sffamily\bfseries\fontsize{5.6}{6}\selectfont] at (12.6,4.8) {candidates in each round};
\node[ln, text=black!55, font=\sffamily\fontsize{5}{6}\selectfont] at (11.6,4.52) {iteration};
\foreach \r/\k/\ca/\cb/\cc in {1/c/stPair/stPred/stBox, 2/f/stPair/stPred/stBox, 3/f/stBox/stPair/stBox, 4/m/stPred/stBox/stBox} {
  \pgfmathsetmacro{\yy}{4.6-\r*0.33}
  \node[ln, text=black!65, font=\sffamily\bfseries\fontsize{5.6}{6}\selectfont] at (11.6,\yy) {\r};
  \pgDoc{12.3}{\yy}{0.5}{c}{\ca}\pgDoc{12.92}{\yy}{0.5}{c}{\cb}\pgDoc{13.54}{\yy}{0.5}{\k}{\cc}
}
\pgDoc{11.43}{3.03}{0.4}{c}{stPair}\pgDoc{11.485}{3.005}{0.4}{c}{stPred}%
\foreach \x/\k/\w/\q in {11.54/c/cause/stBox, 12.39/f/free/stPair, 13.07/m/merge/stPair} {
  \pgDoc{\x}{2.98}{0.4}{\k}{\q}
  \node[ln, anchor=base west, text=black!65, font=\sffamily\fontsize{5.2}{6}\selectfont] at (\x+0.14,2.93) {\w};
}
\draw[flow] (14.0,3.55) -- (14.68,3.55);%
\tzEval{15.02}{3.6}{0.72}
\node[num] at (15.84,3.83) {3};
\node[fmnm, text=verify!80!black, font=\sffamily\bfseries\small] at (15.02,3.05) {evaluate};
\draw[flow] (15.5,3.55) -- (16.18,3.55);
\fmBest{best}{16.72}{3.55}
\node[fmnm, text=rohared, font=\sffamily\bfseries\scriptsize, align=center] at (16.72,3.05) {best\\[-1pt]harness};
\node[spt, text=search!90!black] at (14.3,5.5) {check each candidate};
\foreach \n/\x in {gr/14.75, gk/15.73, gp/16.72} \node[sp=search] (\n) at (\x,4.82) {};
\node[spl, anchor=base] at (14.75,5.12) {review};
\node[spl, anchor=base] at (15.73,5.12) {checks};
\node[spl, anchor=base] at (16.72,5.12) {pass};
\begin{scope}[shift={(14.75,4.82)}, scale=0.72]\draw[search!85, line width=0.7pt] (-0.03,0.04) circle (0.1); \draw[search!85, line width=1.1pt, line cap=round] (0.05,-0.04) -- (0.13,-0.12);\end{scope}
\begin{scope}[shift={(15.73,4.82)}, scale=0.72]%
  \fill[search!85] (-0.17,-0.13) rectangle (-0.12,0.14);
  \draw[search!85, line width=0.4pt, fill=white] (-0.12,0.02) rectangle (0.18,0.1);
  \foreach \i in {0,1,2} \fill[diag!70] (-0.08+\i*0.1,0.02) rectangle ++(0.05,0.08);
  \fill[search!85] (0.14,-0.13) rectangle (0.17,0.02);
\end{scope}
\begin{scope}[shift={(16.72,4.82)}, scale=0.72]\draw[search!85, line width=1.0pt, line cap=round, line join=round] (-0.13,0) -- (-0.04,-0.1) -- (0.14,0.11);\end{scope}
\draw[->, line width=0.6pt, search!85, >={Stealth[length=1.5mm]}] (gr.east) -- (gk.west);
\draw[->, line width=0.6pt, search!85, >={Stealth[length=1.5mm]}] (gk.east) -- (gp.west);
\draw[->, line width=0.6pt, search!75, >={Stealth[length=1.4mm]}] (gk.south west) .. controls (15.33,4.3) and (14.7,4.3) .. (gr.south);
\node[spl, text=diag!85, font=\sffamily\fontsize{5.2}{6}\selectfont] at (15.62,4.37) {repair};%
\draw[search!45, line width=0.6pt, densely dotted] (14.2,3.58) -- (14.2,4.82) -- (14.525,4.82);
\newcommand{\fmPill}[4]{%
  \node[fill=#3, rounded corners=0.24cm, minimum height=0.46cm, inner xsep=5pt, font=\sffamily\bfseries\footnotesize, text=white] (pl) at (#1,#2) {\hspace{0.36cm}#4};
  \draw[white, line width=1.1pt, -{Stealth[length=1.3mm, width=1.6mm]}] ($(pl.west)+(0.42,0.0)$) arc[start angle=0, end angle=285, radius=0.13];}
\draw[->, line width=1.3pt, rounded corners=9pt, search!85!black] (15.02,2.68) -- (15.02,2.3) -- (8.35,2.3) -- (8.35,2.75);
\fmPill{11.55}{2.3}{search!85!black}{$\times$4 iterations}
\node[ln, text=search!95!black] at (11.55,1.9) {reads earlier code, scores, records};
\draw[->, line width=1.3pt, rounded corners=10pt, diag!85!black] (16.72,2.5) -- (16.72,1.42) -- (6.55,1.42) -- (6.55,2.75);
\fmPill{11.63}{1.42}{diag!85!black}{$\times$3 rounds}%
\node[ln, text=diag!90!black] at (11.63,1.03) {the RCA agent re-diagnoses the best harness};
\draw[->, line width=1.3pt, rohared] (17.2,3.55) -- (17.85,3.55);
\node[num] at (17.545,3.83) {4};
\fmImage{18.35}{5.3}{1.45}
\node[ln, text=black!70, font=\sffamily\fontsize{6}{7}\selectfont] at (18.35,4.78) {image};
\draw[->, line width=1.1pt, rohared!60] (18.35,4.62) -- (18.35,4.1);
\node[draw=rohared!90!black, line width=1.3pt, fill=rohared!10, rounded corners=4pt, minimum width=0.95cm, minimum height=0.9cm, inner sep=0pt, drop shadow={opacity=0.15, shadow xshift=0.7pt, shadow yshift=-0.7pt}] (dh) at (18.35,3.55) {};
\node[font=\ttfamily\bfseries\fontsize{7}{7}\selectfont, text=rohared, inner sep=0pt] at (18.35,3.72) {</>};
\foreach \i/\c in {0/stBox,1/stName,2/stPair,3/stPred,4/stRank} \fill[\c] (18.35-0.355+\i*0.15,3.27) rectangle ++(0.11,0.11);
\tzLock{18.72}{3.92}{0.9}
\node[ln, text=rohared, font=\sffamily\bfseries\fontsize{6}{7}\selectfont] at (18.35,2.82) {deployed\\[-1pt]harness};
\draw[->, line width=1.1pt, rohared!60] (18.35,2.5) -- (18.35,1.98);
\fmGraph{18.35}{1.66}{1.55}
\begin{scope}[shift={(18.35,1.66)}, scale=1.55]%
  \draw[rohared, line width=1.0pt, -{Stealth[length=1.2mm]}] (0.27,0.1) -- (0.33,-0.17);
  \fill[rohared] (0.35,-0.24) circle (0.07);
\end{scope}
\node[ln, text=black!70, font=\sffamily\fontsize{6}{7}\selectfont, align=center] at (18.35,0.98) {improved\\[-1pt]scene graph};
\end{tikzpicture}}
\caption{\textbf{\methodname{} overview.} (1) The RCA agent corrects components of the frozen model $M$'s output with ground truth and splits the missed ground-truth triplets over the components. It tests each cause with an intervention. (2) In each iteration, the proposer agent writes three candidate harnesses, at least two of which target diagnosed causes. (3) Each is reviewed, checked and scored on \searchsplit images. The best harness is re-diagnosed every four iterations. (4) The harness selected on \interimsplit{} images is deployed around the unchanged model.}
\label{fig:method}
\end{figure*}

\section{Related Work}
\label{sec:related}

\paragraph{Scene graph generation.}
SGG models include two-stage methods that detect objects first and then predict their predicates~\cite{zellers2018motifs,zheng2023penet,neau2026reactpp}, one-stage methods that predict objects and predicates together with a DETR-style decoder~\cite{li2022sgtr,im2024egtr}, and multimodal LLMs that write the graph as text~\cite{chen2025r1sgg,sggr32026}.
Most existing methods improve these models by training them, for example with predicate debiasing and long-tail learning~\cite{tang2020tde,zhang2022ietrans,qu2026saliencesgg}, open-vocabulary training~\cite{chen2024ovsgtr} or reinforcement learning for MLLMs~\cite{chen2025r1sgg,sggr32026}. This training must be repeated for every model and dataset.
Training-free methods post-process the output, for example by filtering predicates with an LLM or VLM~\cite{jiang2025hiercom}, blending debiased VLM scores~\cite{wang2024predicate}, choosing from conformal prediction sets~\cite{nag2025pcsgg} or refining the ranking with commonsense constraints~\cite{kgref2026}.
Each adds one post-processing step that corrects the predicates or the ranking, and applies it in the same way to every model. Without retraining the SGG model, we instead search over the whole inference code around it, so that the fix targets the components where each model fails on each dataset.

\paragraph{Error and root-cause analysis.}
In computer vision, error diagnosis tools have been built for object detection~\cite{hoiem2012diagnosing,bolya2020tide}, temporal action detection~\cite{alwassel2018diagnosing} and human-object interaction detection~\cite{zhu2023diaghoi}. They fix each error type with an oracle and measure how much the score rises, as TIDE~\cite{bolya2020tide} does for false positives, missed objects and localization errors of detectors.
For SGG, earlier analyses each cover part of where ground-truth triplets are lost, such as pair recall~\cite{wang2023pairnet}, detector reachability~\cite{qu2026reachability}, metrics that separate object, pair and predicate quality~\cite{lorenz2024metrics} and ground-truth substitution in a fixed order~\cite{han2021benchmark}.
Outside vision, related methods identify which agent caused a failure in LLM multi-agent systems~\cite{zhang2025whoandwhen}, analyze the root causes of cloud incidents with LLM agents~\cite{wang2024rcagent}, and diagnose approximate nearest-neighbor search to redesign it automatically~\cite{zhou2026smartann}.
When ground truth is substituted in a fixed order, the measured contribution of each component depends on that order~\cite{bolya2020tide}, and the module most responsible for a failure is not always the best place to fix it~\cite{yoon2026diagnosis}.
We measure all five components for every model and dataset with one protocol and split the gain of fixing them with order-independent Shapley values~\cite{shapley1953value}, as ShapleyFlow~\cite{yang2026shapleyflow} does for the modules of agentic workflows. We then test each suspected cause with an intervention whose effect is predicted before it runs.

\paragraph{Agentic harness optimization.}
LLMs can compose vision models and tools into programs~\cite{gupta2023visprog,suris2023vipergpt}.
Such programs are usually designed manually, as in a training-free pipeline that detects human-object interactions with an MLLM~\cite{cai2026harnesshoi}.
Instead of manual design, other work lets LLMs optimize prompts and LM pipelines~\cite{khattab2024dspy,yuksekgonul2025textgrad,agrawal2026gepa}, programs~\cite{romeraparedes2024funsearch} and agent designs~\cite{hu2025adas,zhang2025aflow} automatically.
Meta-Harness~\cite{lee2026metaharness} extends this to the whole harness of an LLM system, and concurrent work applies harness optimization to video understanding~\cite{lin2026videoharness,cui2026metavideoagent}, vision-tool pipelines~\cite{ge2026vtos} and embodied agents~\cite{lee2026emho}.
These optimizers mostly find what to fix by reading the execution traces of an LLM agent, such as its prompts, tool calls and reasoning~\cite{lee2026metaharness,chen2026harnessfix,park2026autosaddler,wang2026t2a}.
An SGG model outputs only the scene graph, without such traces.
We extend agentic harness optimization to SGG and guide the search with root causes measured for each component.

\section{Method}
\label{sec:method}

\methodname has two LLM agents (\cref{fig:method}).
The RCA agent measures where a frozen SGG model loses ground-truth triplets and tests the suspected causes with interventions (\cref{sec:rca}).
The proposer agent then searches for harnesses that fix these causes first (\cref{sec:search}).

\subsection{Problem Setup}
\label{sec:setup-problem}
Let $M$ be a frozen SGG model.
A \emph{harness} $h$ is the inference code around $M$.
Given an image, $h$ prepares the input of $M$, calls $M$, runs external tools, and post-processes the output into a ranked list of (subject, predicate, object) triplets with boxes.
The \emph{baseline} $h_0$ is the official inference pipeline released with $M$ (preprocessing, decoding, post-processing).
With $M$ frozen and a fixed number of search iterations, harness optimization finds the harness that maximizes a score $\mu$ on the data distribution $\mathcal{D}$,
\begin{equation}
h^\star=\operatorname*{arg\,max}_{h\in\mathcal{H}_c}\ \mu\big(h(\cdot\,;M),\,\mathcal{D}\big),
\label{eq:objective}
\end{equation}
where $\mathcal{H}_c$ is the set of harnesses allowed in the inference condition $c$ (single-pass or multi-pass).
The score $\mu$ is the sum of \gc R@100 and mR@100, so it rewards gains on both frequent and rare predicates. The candidate ranking (\cref{sec:search}) also checks $K{=}20$ and 50.
The search approximates $\mathcal{D}$ with the \searchsplit split, selects on the \interimsplit split and scores the selected harness once on \testsplit.

The two inference conditions differ in how many times a harness may call $M$.
A \textbf{single-pass} harness calls $M$ once per image on the unchanged image, as the baseline does, so its gains do not come from extra model calls.
A \textbf{multi-pass} harness itself decides how many times to call $M$ (\eg on crops or flipped images), as long as its estimated time per image stays within 30\,s.

\begin{figure}[t]
\centering
\resizebox{\columnwidth}{!}{%
\begin{tikzpicture}
\input{figures/generated/lossmap_k20_final}
\end{tikzpicture}}
\caption{\textbf{Where the baselines miss ground-truth triplets.} Each cell gives the \gc R@20 or mR@20 points that the baseline loses at that component, \ie its Shapley share (\cref{eq:shapley}) of the gain of correcting all components with ground truth. A triplet that needs several corrections is split equally among them. VG150 uses \numImagesLossMapRoneSGGVGSearch{} \searchsplit and check images, and PSG uses the \numImagesLossMapRoneSGGPSGSearch{} \searchsplit images with ground-truth triplets. All cells share one color scale, and bold marks the largest value per row. Each row sums to the gain of correcting all components, within about 1 point of $100-\text{R@20}$ for \gc R@20.}
\label{fig:lossmap}
\end{figure}

\subsection{Root-Cause Analysis}
\label{sec:rca}
\paragraph{Errors per component.}
A ground-truth triplet counts as a match at $K$ only if it passes five components, a box with IoU${\geq}0.5$ (\emph{box}) and a correct name (\emph{name}) for both objects, a proposal for the pair (\emph{pair}), the correct predicate (\emph{predicate}) and a rank within the top $K$ (\emph{rank}).
To measure how many triplets a harness loses at each component, we correct its output with ground truth.
Let $\mathcal{T}$ be the set of the five components. For a subset $S\subseteq\mathcal{T}$, $v(S)$ is the \gc R@$K$ or mR@$K$ of the harness after the components in $S$ are corrected, so $v(\emptyset)$ is its unchanged recall.
For example, $v(\{\text{box}\})$ adds the missed ground-truth boxes and keeps everything else as predicted.
Following TIDE~\cite{bolya2020tide}, one can correct a single component $s$ and measure how much the recall rises, $\Delta_s = v(\{s\}) - v(\emptyset)$.
This misses triplets that are lost at more than one component.
For example, a triplet whose subject has no box is recovered only when both its box and its pair are corrected. Neither $\Delta_{\text{box}}$ nor $\Delta_{\text{pair}}$ counts it, so the single-component gains do not add up to the missed recall.
Correcting the components one after another in a fixed order~\cite{han2021benchmark} counts it, but attributes it entirely to the component corrected last, so the result depends on the order.
The five components can be corrected in $5!=120$ orders, and on the images of the \searchsplit split the component with the largest share changes in up to half of them. For example, in pipeline order (box, name, pair, predicate, rank), the box component receives nothing for any model. Correcting boxes alone recovers no ground-truth triplet.
We therefore split the gain of correcting all components, $v(\mathcal{T})-v(\emptyset)$, with Shapley values~\cite{shapley1953value}, which average the gain of adding $s$ over all 120 orders.
In each order, $s$ is added after some subset $S$ of the other components, so the average can be computed exactly from the $2^5=32$ subsets of $\mathcal{T}$,
\begin{equation}
\phi_s=\sum_{\makebox[0pt]{$\scriptstyle S\subseteq\mathcal{T}\setminus\{s\}$}}\ \frac{|S|!\,(n-|S|-1)!}{n!}\big(v(S\cup\{s\})-v(S)\big),
\label{eq:shapley}
\end{equation}
where $n{=}|\mathcal{T}|{=}5$ and the weight is the share of the $n!$ orders in which exactly the components in $S$ come before $s$.
For a single triplet, this splits its recovery equally among the components it needs, so the triplet in the example above counts one half for box and one half for pair.
The $\phi_s$ therefore sum to $v(\mathcal{T})-v(\emptyset)$ and do not depend on any order.
For R@$K$, $v(\mathcal{T})-v(\emptyset)$ is within about 1 point of the missed recall $100-\text{R@}K$, so $\phi_s$ reads as the recall points lost at component $s$ (\cref{fig:lossmap}).

\paragraph{RCA agent.}
From these errors per component, per-image records and a literature guide, the RCA agent writes candidate root causes, each a hypothesis about why triplets are lost at one component.
It tests each cause with an intervention (\eg adding boxes from an external detector).
Before each intervention, it writes down the expected direction and size of the change in the metric, so that a cause is accepted only when the predicted change appears.
A cause is \emph{confirmed} if the change has the predicted direction and at least half the predicted size, its paired bootstrap 95\% interval on the \searchsplit images excludes zero, and it has the same sign on the check set, a small image set disjoint from the \searchsplit split (\cref{sec:setup}). Otherwise, it is kept as an \emph{estimated} cause.
For each cause, the agent records the component, image evidence, confirmation experiment and expected direction of the fix.
It then ranks the causes and states a reason for each rank (\cref{fig:method}).
Another LLM agent then reviews the whole analysis. It can send the analysis back to the RCA agent for correction at most twice.
The RCA agent reads only records of the \searchsplit images and the check set, and never sees the \interimsplit or \testsplit images.
After each round $r$, it analyzes the new best harness $h_r$ in the same way, and the next round works on these new causes.

\subsection{Harness Search}
\label{sec:search}
\begin{table}[t]\centering\footnotesize\setlength{\tabcolsep}{3pt}
\caption{\textbf{Harness space by component.} What a harness may change at each component, and the tools and cues it may use. The unguided, knowledge-guided and RCA-guided searches (\cref{sec:levels}) share this space. For R1-SGG, the harness may also optimize prompts with GEPA~\cite{agrawal2026gepa}.}
\label{tab:space}
\begin{tabular*}{\columnwidth}{@{\extracolsep{\fill}}l >{\raggedright\arraybackslash}p{2.75cm} >{\raggedright\arraybackslash}p{3.4cm}@{}}\toprule
Component & Harness operations & Tools and cues\\\midrule
box & add, refine or merge boxes & Grounding DINO~\cite{liu2024groundingdino}, OWLv2~\cite{minderer2023owlv2}, SAM~2.1~\cite{ravi2025sam2}\\
name & correct or add object names & RAM++~\cite{huang2023rampp}, Florence-2~\cite{xiao2024florence2}, CLIP~\cite{radford2021clip}, DINOv2~\cite{oquab2024dinov2}\\
pair & rebuild candidate pairs & box geometry, detector outputs\\
predicate & rescore predicates with evidence & RTMPose~\cite{jiang2023rtmpose}, Florence-2 captions, Depth Anything~V2~\cite{yang2024depthanything}\\
rank & re-rank the triplet list & model scores, priors from training images\\\bottomrule
\end{tabular*}\end{table}

\begin{table*}[t]
\settowidth{\dupw}{\scriptsize(+26.40)}%
\colorlet{rohatint}{black!4}%
\centering\small\setlength{\tabcolsep}{3.5pt}
\caption{\textbf{Main results on the official \testsplit sets.} All metrics are graph-constrained (\gc) per-image means. Parentheses give the change against the baseline, computed before rounding. Bold and underline mark the best and second best per model and dataset. Base networks are listed under the models. The bottom rows are published numbers under the authors' protocols. $^\dagger$R1-SGG's own protocol, with several predicates per pair and VG150 on its validation split. $^\ddagger$Public model card. The R1-SGG baseline writes at most 20 triplets for almost every image, so its R@20, R@50 and R@100 nearly coincide.}
\label{tab:main}
\resizebox{\textwidth}{!}{%
\begin{tabular}{ll l cccccc c c}
\toprule
Data & Model & Harness & R@20 & R@50 & R@100 & mR@20 & mR@50 & mR@100 & F@100 & AP50\\
\midrule
VG150 & REACT++ & baseline & 23.39\dnone & 30.12\dnone & 34.01\dnone & 12.25\dnone & 15.96\dnone & 18.15\dnone & 23.67\dnone & 37.25\dnone\\
\rowcolor{rohatint} & {\scriptsize\color{black!60}YOLOv8m} & single-pass~{\scriptsize(ours)} & \underline{24.21}\,\dup{+0.82} & \textbf{32.09}\,\dup{+1.97} & \textbf{37.27}\,\dup{+3.27} & \underline{13.31}\,\dup{+1.06} & \textbf{17.46}\,\dup{+1.50} & \textbf{20.46}\,\dup{+2.30} & \textbf{26.42}\,\dup{+2.75} & \textbf{37.72}\,\dup{+0.47}\\
\rowcolor{rohatint} &  & multi-pass~{\scriptsize(ours)} & \textbf{24.48}\,\dup{+1.09} & \underline{31.82}\,\dup{+1.69} & \underline{36.34}\,\dup{+2.33} & \textbf{13.54}\,\dup{+1.29} & \underline{17.34}\,\dup{+1.38} & \underline{19.87}\,\dup{+1.72} & \underline{25.69}\,\dup{+2.02} & \underline{37.67}\,\dup{+0.42}\\
\cmidrule(lr){2-11}
VG150 & EGTR & baseline & 24.38\dnone & 31.04\dnone & 35.17\dnone & 5.46\dnone & 7.97\dnone & 10.11\dnone & 15.70\dnone & 26.75\dnone\\
\rowcolor{rohatint} & {\scriptsize\color{black!60}ResNet-50} & single-pass~{\scriptsize(ours)} & \textbf{25.75}\,\dup{+1.37} & \underline{32.66}\,\dup{+1.62} & \underline{36.90}\,\dup{+1.74} & \underline{9.19}\,\dup{+3.73} & \underline{14.51}\,\dup{+6.54} & \underline{18.77}\,\dup{+8.66} & \underline{24.88}\,\dup{+9.18} & \underline{26.83}\,\dup{+0.08}\\
\rowcolor{rohatint} &  & multi-pass~{\scriptsize(ours)} & \underline{25.64}\,\dup{+1.27} & \textbf{33.29}\,\dup{+2.25} & \textbf{38.80}\,\dup{+3.63} & \textbf{10.45}\,\dup{+4.98} & \textbf{16.09}\,\dup{+8.11} & \textbf{20.16}\,\dup{+10.05} & \textbf{26.53}\,\dup{+10.83} & \textbf{27.09}\,\dup{+0.35}\\
\cmidrule(lr){2-11}
VG150 & R1-SGG & baseline & 24.06\dnone & 24.12\dnone & 24.12\dnone & 6.33\dnone & 6.35\dnone & 6.35\dnone & 10.05\dnone & 20.53\dnone\\
\rowcolor{rohatint} & {\scriptsize\color{black!60}Qwen2-VL 7B} & single-pass~{\scriptsize(ours)} & \underline{25.25}\,\dup{+1.19} & \underline{30.23}\,\dup{+6.10} & \underline{32.15}\,\dup{+8.03} & \underline{8.95}\,\dup{+2.63} & \underline{10.72}\,\dup{+4.37} & \underline{13.55}\,\dup{+7.20} & \underline{19.06}\,\dup{+9.01} & \underline{25.64}\,\dup{+5.11}\\
\rowcolor{rohatint} &  & multi-pass~{\scriptsize(ours)} & \textbf{28.20}\,\dup{+4.15} & \textbf{35.05}\,\dup{+10.93} & \textbf{38.67}\,\dup{+14.55} & \textbf{10.78}\,\dup{+4.45} & \textbf{13.76}\,\dup{+7.41} & \textbf{15.18}\,\dup{+8.83} & \textbf{21.80}\,\dup{+11.75} & \textbf{26.40}\,\dup{+5.86}\\
\midrule
PSG & REACT++ & baseline & 30.46\dnone & 35.91\dnone & 38.99\dnone & 22.53\dnone & 25.25\dnone & 26.93\dnone & 31.85\dnone & 55.22\dnone\\
\rowcolor{rohatint} & {\scriptsize\color{black!60}YOLOv8m} & single-pass~{\scriptsize(ours)} & \underline{31.03}\,\dup{+0.57} & \textbf{37.43}\,\dup{+1.52} & \textbf{40.63}\,\dup{+1.64} & \underline{23.26}\,\dup{+0.73} & \underline{25.35}\,\dup{+0.10} & \underline{27.59}\,\dup{+0.66} & \underline{32.86}\,\dup{+1.01} & \textbf{55.54}\,\dup{+0.31}\\
\rowcolor{rohatint} &  & multi-pass~{\scriptsize(ours)} & \textbf{31.60}\,\dup{+1.14} & \underline{37.29}\,\dup{+1.38} & \underline{40.17}\,\dup{+1.18} & \textbf{24.49}\,\dup{+1.96} & \textbf{27.58}\,\dup{+2.33} & \textbf{29.52}\,\dup{+2.60} & \textbf{34.03}\,\dup{+2.18} & \underline{55.53}\,\dup{+0.31}\\
\cmidrule(lr){2-11}
PSG & R1-SGG & baseline & 28.29\dnone & 28.35\dnone & 28.35\dnone & 11.52\dnone & 11.53\dnone & 11.53\dnone & 16.39\dnone & 42.73\dnone\\
\rowcolor{rohatint} & {\scriptsize\color{black!60}Qwen2-VL 7B} & single-pass~{\scriptsize(ours)} & \underline{32.87}\,\dup{+4.57} & \underline{36.68}\,\dup{+8.33} & \underline{38.61}\,\dup{+10.26} & \underline{28.01}\,\dup{+16.49} & \underline{30.55}\,\dup{+19.02} & \underline{31.88}\,\dup{+20.35} & \underline{34.93}\,\dup{+18.53} & \underline{51.26}\,\dup{+8.53}\\
\rowcolor{rohatint} &  & multi-pass~{\scriptsize(ours)} & \textbf{36.77}\,\dup{+8.48} & \textbf{41.48}\,\dup{+13.13} & \textbf{44.19}\,\dup{+15.84} & \textbf{34.52}\,\dup{+23.00} & \textbf{36.62}\,\dup{+25.09} & \textbf{37.93}\,\dup{+26.40} & \textbf{40.82}\,\dup{+24.43} & \textbf{51.94}\,\dup{+9.21}\\
\midrule
VG150 & REACT++ & card$^\ddagger$ & --\dnone & 29.96\dnone & 34.09\dnone & --\dnone & 16.31\dnone & 18.45\dnone & --\dnone & --\dnone\\
VG150 & EGTR & paper & --\dnone & 30.2\dnone & 34.3\dnone & --\dnone & 7.9\dnone & 10.1\dnone & --\dnone & --\dnone\\
VG150 & R1-SGG & paper$^\dagger$ & \multicolumn{3}{c}{23.75 (R, val)} & \multicolumn{3}{c}{11.43 (mR, val)} & --\dnone & --\dnone\\
PSG & REACT++ & paper & 29.1\dnone & 34.2\dnone & 37.5\dnone & 22.2\dnone & 24.9\dnone & 27.1\dnone & --\dnone & --\dnone\\
PSG & R1-SGG & paper$^\dagger$ & --\dnone & --\dnone & 43.48\dnone & --\dnone & --\dnone & 33.71\dnone & --\dnone & --\dnone\\
\bottomrule
\end{tabular}}
\end{table*}

\paragraph{Harness space.}
Each candidate harness may change any of the five components (\cref{tab:space}).
It may call the external tools listed in \cref{tab:space}, but no other SGG model or SGG ensemble.
It may also train small add-on modules on training images while $M$ stays frozen, for example a calibration that rescales the model's predicate scores or a light head, a small classifier on the model's features.

\paragraph{Proposer agent.}
In each iteration, the proposer agent starts a new session and writes three candidate harnesses.
Following Meta-Harness~\cite{lee2026metaharness}, it reads the code, scores, errors per component and per-image records of all earlier candidates in its search through a file system.
A \emph{cause-guided} candidate receives one diagnosed cause from the RCA agent and writes a harness that fixes it.
It starts from a parent harness, which is an earlier candidate of the search.
In the first iteration of a round, all three candidates are cause-guided. In the next two iterations, two are cause-guided, and the third is a \emph{free} candidate. A free candidate receives no cause, so it can explore any change to the harness.
In the last iteration, two candidates are again cause-guided, and the third merges the fixes found by the round's cause-guided candidates into one harness.
A cause is not given again once its error is fixed or after two attempts that do not reduce it.
Before scoring, a reviewer agent, separate from the proposer, checks each candidate's code. Automatic checks also ensure that the harness follows its inference condition, for example that a single-pass harness calls $M$ only once and that no harness reads ground truth.

\paragraph{Ranking candidate harnesses.}
For each candidate harness, we measure its gain over the baseline $h_0$ on the six primary metrics (\gc R@$K$ and mR@$K$, $K\in\{20,50,100\}$) on the \searchsplit split.
A candidate ranks higher when more of these gains have an image-paired bootstrap 95\% interval above zero.
Among candidates with the same count, the one with the larger gain in the score $\mu$ (\cref{eq:objective}) ranks higher.
If two gains in $\mu$ differ by less than 0.3 points, the gain at $K{=}50$ decides, and then the gain at $K{=}20$.
The free candidates of the proposer also use this ranking to choose their parent harness. Among earlier candidates other than the current best, they pick the one with the highest ranking score plus an exploration bonus, which shrinks the more often that candidate has already served as a parent (an upper-confidence rule).

\paragraph{Rounds and final selection.}
Each search runs three rounds of four iterations, 12 iterations in total.
A candidate harness is promoted if it raises at least one primary metric on the \searchsplit split and no other primary metric drops significantly.
Among the promoted candidates, the main harness of each search is chosen by the same ranking on the \interimsplit split. No agent sees these images during the search, so the choice does not depend on the images that guided it.

\section{Experiments}
\label{sec:experiments}

\subsection{Experimental Setup}
\label{sec:setup}

\paragraph{Models and data.}
We use the publicly released weights of one model per architecture, the two-stage REACT++~\cite{neau2026reactpp}, the one-stage EGTR~\cite{im2024egtr} and the multimodal LLM R1-SGG~\cite{chen2025r1sgg} (base networks in \cref{tab:main}).
No PSG checkpoint is released for EGTR, so VG150~\cite{xu2017imp,krishna2017vg} and PSG~\cite{yang2022psg} give five model--dataset combinations.
For each combination, we run an RCA-guided search under the single-pass and the multi-pass condition (\cref{sec:setup-problem}).
To measure the effect of root-cause analysis, the VG150 single-pass searches of the three models are also run as unguided and knowledge-guided searches (\cref{sec:levels}).
All agents use Claude Opus 5.5 at maximum reasoning effort.

\paragraph{Splits.}
The \textbf{\searchsplit} split scores every candidate harness, and the agents see its per-image records.
The \textbf{\interimsplit} split is hidden from all agents. After the search, it re-scores the promoted candidates to pick the main harness among them (\cref{sec:search}).
The \textbf{\testsplit} split is the official test set.
A small check set lets the RCA agent repeat each confirmation experiment on images outside the \searchsplit split (\cref{sec:rca}).
On VG150, the \searchsplit (2{,}500 images), check (500) and \interimsplit (1{,}000) splits are disjoint subsets of the official validation set and contain no image from the training set of any VG150 model. PSG has no official validation set, so its \searchsplit (791), check (150) and \interimsplit (997) splits come from its official training set.
We use all 26{,}446 VG150 test images and all 2{,}177 PSG test images with ground-truth triplets.

\paragraph{Metrics.}
Following the standard SGG evaluation protocol~\cite{tang2020sggcode,chen2025r1sgg}, we report graph-constrained (\gc, one predicate per subject--object pair) R@$K$ and mR@$K$ at $K\in\{20,50,100\}$, averaged over images.
For a fair comparison, we score all models with the same evaluation code.
F@100 is the harmonic mean of R@100 and mR@100~\cite{zhang2022ietrans}.
On PSG, a predicted box matches a ground-truth box if their IoU is at least 0.5 and their labels agree, as in the evaluation of REACT++ and R1-SGG.
We also report detection AP50~\cite{lin2014coco}.
All intervals are 95\% intervals from an image-paired bootstrap.

\begin{table}[t]\centering\footnotesize\setlength{\tabcolsep}{3pt}
\caption{\textbf{Inference time per image.} Wall-clock time of the model's forward pass (baseline) and, for the selected harness, of that pass plus its extra model calls, external tools and harness computation. Times are medians, measured one image at a time on one RTX~PRO~6000. External detectors are not speed-optimized.}\label{tab:time}
\begin{tabular*}{\columnwidth}{@{\extracolsep{\fill}}llccc@{}}\toprule
 & & \multicolumn{3}{c}{Inference time (s per image)}\\\cmidrule(l){3-5}
Data & Model & Baseline & Single-pass & Multi-pass\\\midrule
VG150 & REACT++ & 0.017 & 0.92 & 1.27\\
 & EGTR & 0.075 & 0.076 & 2.27\\
 & R1-SGG & 11.7 & 13.2 & 29.1\\
\midrule
PSG & REACT++ & 0.015 & 0.75 & 1.01\\
 & R1-SGG & 11.0 & 12.5 & 29.7\\
\bottomrule\end{tabular*}\end{table}

\subsection{Main Results}
\label{sec:main}
\begin{figure}[t]
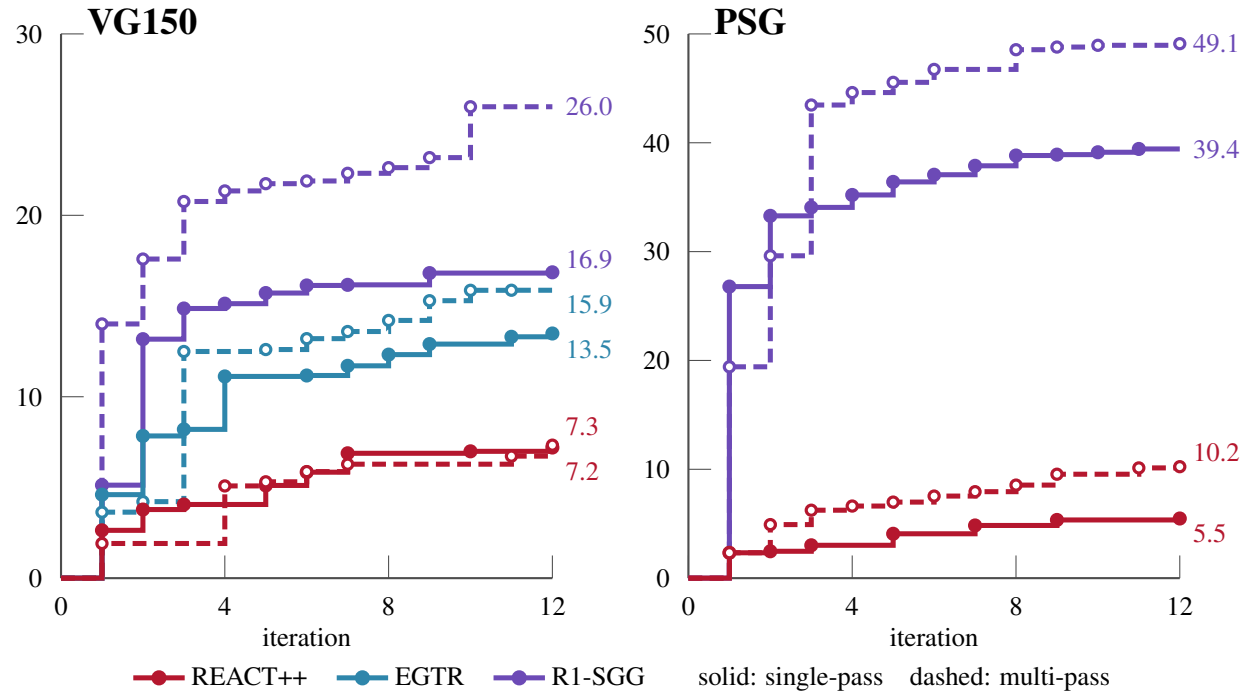

\centering
\colorlet{cbBlue}{stPair}\colorlet{cbGreen}{stPred}\colorlet{cbVerm}{rohared}%
\pgfplotsset{
  mcR/.style={cbBlue, line width=0.95pt, solid, mark=*, mark size=0.85pt},
  mcRM/.style={cbBlue, line width=0.95pt, dash pattern=on 2.6pt off 1.4pt, mark=*, mark size=0.95pt, mark options={solid, fill=white, line width=0.6pt}},
  mcE/.style={cbGreen, line width=0.95pt, solid, mark=*, mark size=0.85pt},
  mcEM/.style={cbGreen, line width=0.95pt, dash pattern=on 2.6pt off 1.4pt, mark=*, mark size=0.95pt, mark options={solid, fill=white, line width=0.6pt}},
  mcP/.style={cbVerm, line width=0.95pt, solid, mark=*, mark size=0.85pt},
  mcPM/.style={cbVerm, line width=0.95pt, dash pattern=on 2.6pt off 1.4pt, mark=*, mark size=0.95pt, mark options={solid, fill=white, line width=0.6pt}}}
\tikzset{mcRLab/.style={text=cbBlue, font=\tiny\bfseries}, mcRMLab/.style={text=cbBlue}, mcELab/.style={text=cbGreen}, mcEMLab/.style={text=cbGreen},
  mcPLab/.style={text=cbVerm}, mcPMLab/.style={text=cbVerm}}
\pgfplotsset{every tick label/.append style={font=\rmfamily}, /pgf/number format/assume math mode=true}%
\begin{tikzpicture}
\begin{groupplot}[group style={group size=2 by 1, horizontal sep=0.9cm},
  scale only axis, width=3.25cm, height=3.6cm, clip=false, xmin=0, xmax=36, ymin=0, xtick={0,12,24,36}, xticklabels={0,4,8,12},
  x tick label style={font=\tiny}, y tick label style={font=\tiny}, title style={font=\scriptsize\bfseries, at={(rel axis cs:0.04,1)}, anchor=north west, yshift=0pt, inner sep=1pt},
  xlabel={\tiny iteration}, xlabel style={yshift=1.0ex},
  axis x line*=bottom, axis y line*=left, axis line style={black!70}, tick style={black!70}]
\nextgroupplot[title={VG150}, ymax=30, ytick={0,10,20,30}]
\input{figures/generated/main_curves_small_VG_sq}
\nextgroupplot[title={PSG}, ymax=50, ytick={0,10,20,30,40,50}]
\input{figures/generated/main_curves_small_PSG}
\end{groupplot}
\node[anchor=north, font=\tiny, inner sep=0pt] at ($(group c1r1.south)!0.5!(group c2r1.south)+(0,-0.58)$) {%
\tikz[baseline=-0.6ex]\draw[cbVerm, line width=0.95pt] (0,0) -- (0.34,0) plot[mark=*, mark size=0.95pt] coordinates {(0.17,0)};\,REACT++\quad
\tikz[baseline=-0.6ex]\draw[cbGreen, line width=0.95pt] (0,0) -- (0.34,0) plot[mark=*, mark size=0.95pt] coordinates {(0.17,0)};\,EGTR\quad
\tikz[baseline=-0.6ex]\draw[cbBlue, line width=0.95pt] (0,0) -- (0.34,0) plot[mark=*, mark size=0.95pt] coordinates {(0.17,0)};\,R1-SGG\qquad
solid: single-pass\quad dashed: multi-pass};
\end{tikzpicture}
\caption{\textbf{Search progress.} Best harness so far (by the ranking of \cref{sec:search}), as a running maximum, in each of the 10 main searches. All ten still improve in round~3. The $y$-axis shows the gain over the baseline, $\Delta$R@100$+\Delta$mR@100 (\gc, \searchsplit split).}
\label{fig:maincurves-small}
\end{figure}

\Cref{tab:main} compares, for each model--dataset combination, the baseline with the harnesses selected by the single-pass and multi-pass searches.
The bottom rows of \cref{tab:main} list the numbers published for each model.
Our REACT++ and EGTR baselines are within about 1.5 points of them at R@100 and mR@100.
R1-SGG's published numbers differ more because its own evaluation code counts several predicates per subject--object pair. With that code, our R1-SGG baseline gives 43.49 R@100 and 33.75 mR@100 on PSG, against 43.48 and 33.71 in its paper.
\Cref{fig:maincurves-small} shows the best score found so far over the iterations of each search, measured on the \searchsplit split. None of the searches has saturated after 12 iterations.

\begin{figure*}[t]
\centering
\resizebox{\textwidth}{!}{%
\begin{tikzpicture}[font=\sffamily\small,
  on/.style={draw=black!35, fill=black!6, rounded corners=5pt, line width=0.6pt, font=\sffamily\scriptsize, text=black!80, inner xsep=4pt, inner ysep=2.2pt, minimum height=0.42cm},
  ghost/.style={draw=black!45, fill=white, dash pattern=on 2.2pt off 1.4pt, rounded corners=5pt, line width=0.85pt, font=\sffamily\scriptsize, text=black!45, inner xsep=4pt, inner ysep=2.2pt, minimum height=0.42cm},
  rec/.style={-{Stealth[length=2.4mm,width=1.9mm]}, #1, line width=1.6pt, shorten <=1pt, shorten >=1pt},
  oth/.style={-{Stealth[length=1.9mm,width=1.4mm]}, black!65, line width=0.75pt, shorten <=1pt, shorten >=1pt},
  kept/.style={-{Stealth[length=1.9mm,width=1.4mm]}, black!50, line width=0.75pt, shorten <=1pt, shorten >=1pt},
  lrec/.style={font=\sffamily\scriptsize\bfseries, text=#1!85!black, fill=white, inner sep=1.4pt, rounded corners=2pt},
  loth/.style={font=\sffamily\scriptsize, text=black!70, fill=white, inner sep=1.4pt, rounded corners=2pt},
  lkept/.style={font=\sffamily\scriptsize, text=black!60, fill=white, inner sep=1.4pt, rounded corners=2pt},
  note/.style={font=\sffamily\fontsize{5.6}{6.4}\selectfont, inner sep=1pt},
  newb/.style={fill=stBox, text=white, font=\sffamily\fontsize{5.6}{6.4}\selectfont\bfseries, rounded corners=2pt, inner xsep=2pt, inner ysep=1.2pt},
  cause/.style={fill=white, rounded corners=3pt, line width=1.0pt, text width=3.25cm, minimum height=2.25cm, inner sep=0pt, drop shadow={opacity=0.15, shadow xshift=0.6pt, shadow yshift=-0.6pt}},
  gist/.style={fill=panelGrey, draw=edgeGrey, rounded corners=1.5pt, line width=0.45pt, font=\sffamily\fontsize{5.6}{6.4}\selectfont, text=black, text width=2.81cm, align=left, inner sep=2.2pt},
  r100g/.style={fill=#1, text=white, font=\sffamily\tiny\bfseries, rounded corners=2pt, inner xsep=2.5pt, inner ysep=1.3pt, text height=0.9ex, text depth=0.2ex},
  r100/.style={fill=black!78, fill opacity=0.85, text opacity=1, text=white, font=\sffamily\tiny\bfseries, rounded corners=2pt, inner xsep=2.5pt, inner ysep=1.3pt, text height=0.9ex, text depth=0.2ex},
  bigarrow/.style={-{Triangle[length=2.6mm,width=3.4mm]}, line width=2.2pt, black!55}]
\foreach \x/\t in {4.56/Baseline, 12.66/+RoHa-SGG, 18.18/Cause $\rightarrow$ fix}
  \node[font=\sffamily\small\bfseries, text=black, anchor=base] at (\x,1.5) {\t};
\draw[edgeGrey, line width=0.6pt] (0.05,1.36) -- (19.85,1.36);
\foreach \y/\m/\f in {0/REACT++/two-stage, -2.75/EGTR/one-stage, -5.5/R1-SGG/MLLM} {
  \draw[edgeGrey, line width=0.5pt] (0.05,\y-1.36) -- (19.85,\y-1.36);
  \draw[edgeGrey, fill=white, rounded corners=5pt, line width=0.6pt] (0.82,\y-1.22) rectangle (8.30,\y+1.22);
  \draw[edgeGrey, fill=white, rounded corners=5pt, line width=0.6pt] (8.92,\y-1.22) rectangle (16.40,\y+1.22);
  \node[rotate=90, align=center, font=\sffamily\small, text=black] at (0.42,\y) {\textbf{\m}\\[-3pt]{\scriptsize\color{black!60}\f}};
  \draw[bigarrow] (8.35,\y) -- (8.87,\y);
}
\node[inner sep=0pt, anchor=west] (ima) at (0.90,0) {\includegraphics[height=2.3cm]{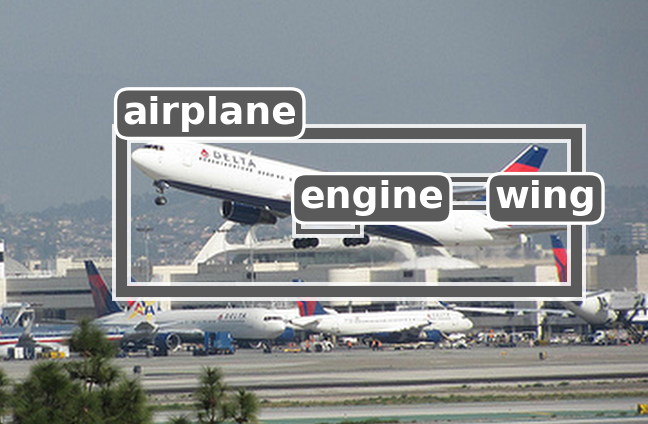}};
\node[r100, anchor=south east] at ($(ima.south east)+(-0.06,0.06)$) {R@100 \,25};
\node[inner sep=0pt, anchor=west] (ime) at (9.00,0) {\includegraphics[height=2.3cm]{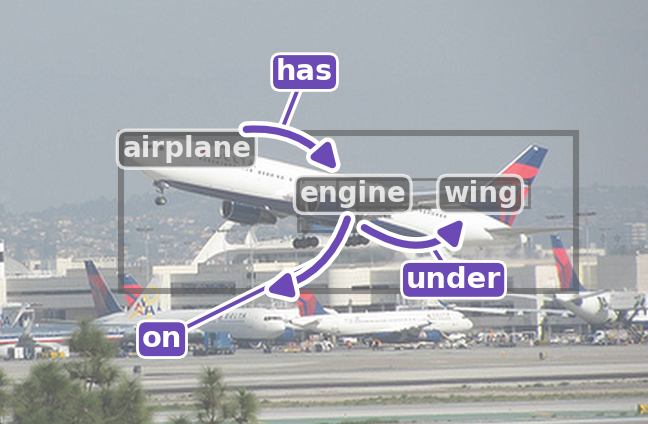}};
\node[r100g=stPair, anchor=south east] (ge) at ($(ime.south east)+(-0.06,0.06)$) {(+75)};
\node[r100, anchor=south east] at ($(ge.south west)+(-0.03,0)$) {R@100 {\mdseries\color{white!80!black} 25\,$\to$}\,100};
\node[on] (b1) at (5.25,0.35) {airplane};
\node[on] (b2) at (7.49,0.35) {engine};
\node[on] (b3) at (7.49,-0.85) {wing};
\node[on] (a1) at (13.36,0.35) {airplane};
\node[on] (a2) at (15.59,0.35) {engine};
\node[on] (a3) at (15.59,-0.85) {wing};
\draw[rec=stPair] (a1) to[bend left=28] node[lrec=stPair] {has} (a2);
\draw[rec=stPair] (a2) to[bend left=28] node[lrec=stPair] {on} (a1);
\draw[rec=stPair] (a2) -- node[lrec=stPair, left] {under} (a3);
\node[cause, draw=stPair, anchor=north west] (k) at (16.55,1.12) {};
\fill[stPair, rounded corners=2pt] ($(k.north west)+(0.14,-0.14)$) rectangle ($(k.north west)+(0.64,-0.50)$);
\tzStage{3}{16.94}{0.800}{1.1}{white}
\node[anchor=north west, font=\sffamily\fontsize{6.5}{7.8}\selectfont\bfseries, text=stPair!90!black, inner sep=0pt] at ($(k.north west)+(0.78,-0.2)$) {pair component};
\node[anchor=north west, font=\sffamily\scriptsize, text=black, text width=2.95cm, align=left, inner sep=0pt] at ($(k.north west)+(0.14,-0.58)$) {Both ends are listed, but\\the list never relates them.};
\node[gist, anchor=south west] at ($(k.south west)+(0.14,0.12)$) {fix: repeat penalty};
\node[inner sep=0pt, anchor=west] (ima) at (0.90,-2.75) {\includegraphics[height=2.3cm]{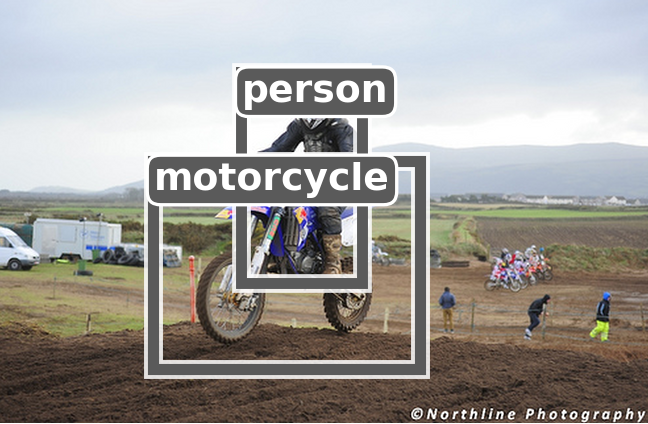}};
\node[r100, anchor=south east] at ($(ima.south east)+(-0.06,0.06)$) {R@100 \,60};
\node[inner sep=0pt, anchor=west] (ime) at (9.00,-2.75) {\includegraphics[height=2.3cm]{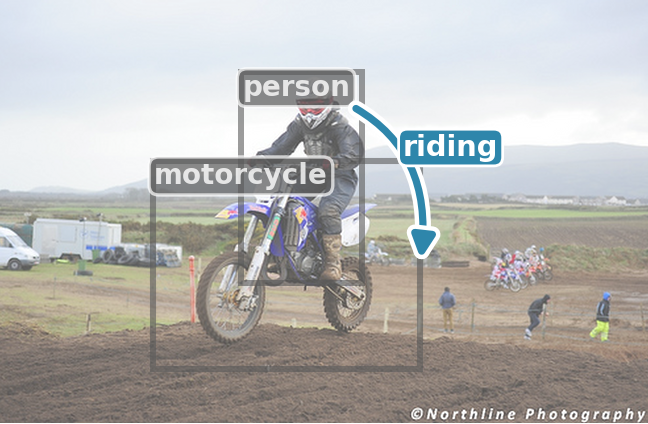}};
\node[r100g=stPred, anchor=south east] (ge) at ($(ime.south east)+(-0.06,0.06)$) {(+20)};
\node[r100, anchor=south east] at ($(ge.south west)+(-0.03,0)$) {R@100 {\mdseries\color{white!80!black} 60\,$\to$}\,80};
\node[on] (b1) at (5.30,-2.13) {person};
\node[on] (b2) at (5.30,-3.37) {motorcycle};
\node[on] (a1) at (13.40,-2.13) {person};
\node[on] (a2) at (13.40,-3.37) {motorcycle};
\draw[kept] (b1) -- node[lkept, right] {on} (b2);
\draw[rec=stPred] (a1) -- node[lrec=stPred, right] {riding} (a2);
\node[on] (b4) at (7.66,-1.98) {helmet};
\node[on] (b5) at (7.66,-2.75) {boot};
\node[on] (b6) at (7.66,-3.52) {bike};
\draw[kept] (b1.east) -- node[lkept, pos=0.5] {wearing} (b4.west);
\draw[kept] (b1.south east) -- node[lkept, pos=0.55] {wearing} (b5.west);
\draw[kept] ($(b1.south)+(0.25,0)$) -- node[lkept, pos=0.72] {riding} (b6.north west);
\node[on] (a4) at (15.77,-1.98) {helmet};
\node[on] (a5) at (15.77,-2.75) {boot};
\node[on] (a6) at (15.77,-3.52) {bike};
\draw[kept] (a1.east) -- node[lkept, pos=0.5] {wearing} (a4.west);
\draw[kept] (a1.south east) -- node[lkept, pos=0.55] {wearing} (a5.west);
\draw[kept] ($(a1.south)+(0.25,0)$) -- node[lkept, pos=0.72] {riding} (a6.north west);
\node[cause, draw=stPred, anchor=north west] (k) at (16.55,-1.63) {};
\fill[stPred, rounded corners=2pt] ($(k.north west)+(0.14,-0.14)$) rectangle ($(k.north west)+(0.64,-0.50)$);
\tzStage{4}{16.94}{-1.950}{1.1}{white}
\node[anchor=north west, font=\sffamily\fontsize{6.5}{7.8}\selectfont\bfseries, text=stPred!90!black, inner sep=0pt] at ($(k.north west)+(0.78,-0.2)$) {predicate component};
\node[anchor=north west, font=\sffamily\scriptsize, text=black, text width=2.95cm, align=left, inner sep=0pt] at ($(k.north west)+(0.14,-0.58)$) {The frequency prior\\collapses predicates\\onto head classes.};
\node[gist, anchor=south west] at ($(k.south west)+(0.14,0.12)$) {fix: debias the predicate prior};
\node[inner sep=0pt, anchor=west] (ima) at (0.90,-5.5) {\includegraphics[height=2.3cm]{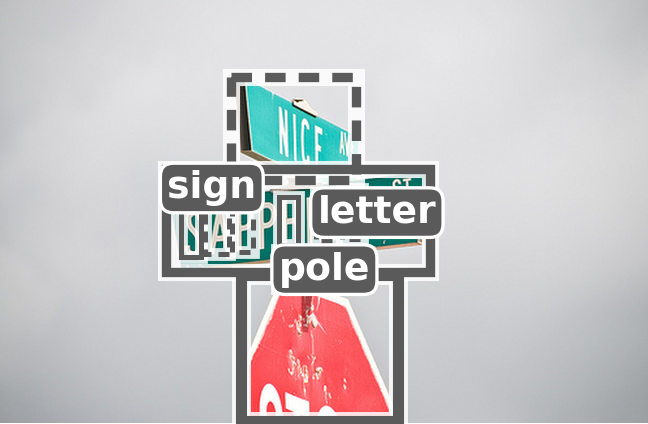}};
\node[r100, anchor=south east] at ($(ima.south east)+(-0.06,0.06)$) {R@100 \,0};
\node[inner sep=0pt, anchor=west] (ime) at (9.00,-5.5) {\includegraphics[height=2.3cm]{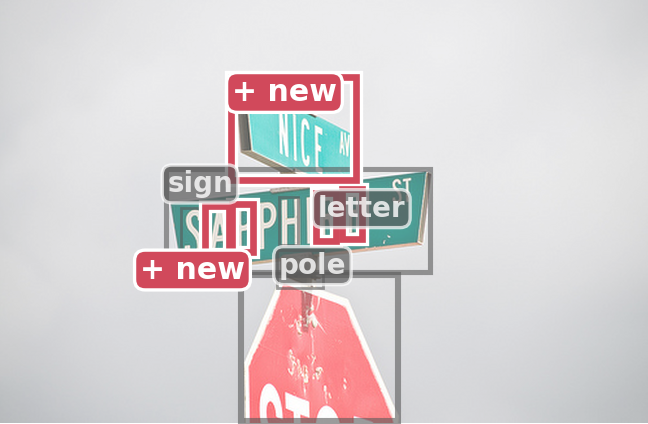}};
\node[r100g=stBox, anchor=south east] (ge) at ($(ime.south east)+(-0.06,0.06)$) {(+100)};
\node[r100, anchor=south east] at ($(ge.south west)+(-0.03,0)$) {R@100 {\mdseries\color{white!80!black} 0\,$\to$}\,100};
\node[ghost] (bL0) at (5.25,-4.88) {letter};
\node[ghost] (bL1) at (5.25,-5.36) {letter};
\node[ghost] (bL2) at (5.25,-5.84) {letter};
\node[on] (bS) at (7.44,-5.36) {sign};
\node[on] (bP) at (7.44,-6.36) {pole};
\node[ghost] (bT) at (5.25,-6.36) {sign};
\node[on, draw=stBox, line width=1.6pt] (aL0) at (13.3,-4.88) {letter};
\node[on, draw=stBox, line width=1.6pt] (aL1) at (13.3,-5.36) {letter};
\node[on, draw=stBox, line width=1.6pt] (aL2) at (13.3,-5.84) {letter};
\node[on] (aS) at (15.55,-5.36) {sign};
\node[on] (aP) at (15.55,-6.36) {pole};
\node[on, draw=stBox, line width=1.6pt] (aT) at (13.3,-6.36) {sign};
\node[note, text=black!60, anchor=south west] at ($(bL0.north west)+(0,0.03)$) {\itshape not boxed};
\node[newb, anchor=south west] at ($(aL0.north west)+(0,0.04)$) {+ new box};
\draw[rec=stBox] (aL0) -- node[lrec=stBox, pos=0.36] {on} (aS.north west);
\draw[rec=stBox] (aL1) -- node[lrec=stBox, pos=0.36] {on} (aS.west);
\draw[rec=stBox] (aL2) -- node[lrec=stBox, pos=0.36] {on} (aS.south west);
\draw[rec=stBox] (aT) -- node[lrec=stBox, pos=0.42] {on} (aP);
\draw[oth] (aS) -- node[loth, right] {on} (aP);
\node[cause, draw=stBox, anchor=north west] (k) at (16.55,-4.38) {};
\fill[stBox, rounded corners=2pt] ($(k.north west)+(0.14,-0.14)$) rectangle ($(k.north west)+(0.64,-0.50)$);
\tzStage{1}{16.94}{-4.700}{1.1}{white}
\node[anchor=north west, font=\sffamily\fontsize{6.5}{7.8}\selectfont\bfseries, text=stBox!90!black, inner sep=0pt] at ($(k.north west)+(0.78,-0.2)$) {box component};
\node[anchor=north west, font=\sffamily\scriptsize, text=black, text width=2.95cm, align=left, inner sep=0pt] at ($(k.north west)+(0.14,-0.58)$) {The model misses\\further instances of\\classes it names.};
\node[gist, anchor=south west] at ($(k.south west)+(0.14,0.12)$) {fix: add detector boxes};
\begin{scope}[shift={(3.2,-7.2)}, lg/.style={font=\sffamily\scriptsize, anchor=west, text=black}]
  \draw[rec=stPair] (0,0.07) -- (0.6,0.07); \draw[rec=stPred] (0,-0.07) -- (0.6,-0.07); \node[lg] (l1) at (0.68,0) {recovered triplet};
  \node[on, fill=white, draw=stBox, line width=1.6pt, minimum height=0.32cm, inner xsep=3pt, anchor=west] (l2) at ($(l1.east)+(0.4,0)$) {obj.}; \node[lg] (l3) at ($(l2.east)+(0.08,0)$) {added box};
  \draw[kept] ($(l3.east)+(0.6,0)$) -- ++(0.6,0) coordinate (l5); \node[lg] (l6) at ($(l5)+(0.08,0)$) {other triplet};
  \node[ghost, minimum height=0.32cm, inner xsep=3pt, anchor=west] (l7) at ($(l6.east)+(0.4,0)$) {obj.}; \node[lg] at ($(l7.east)+(0.08,0)$) {not boxed by the baseline};
\end{scope}
\end{tikzpicture}}
\caption{\textbf{Before and after \methodname, one VG150 \testsplit image per model.} Each image was chosen for clarity among the 30 largest per-image gains with at least two ground-truth triplets recovered at the cause's component. For EGTR, it shows a generic predicate replaced by the annotated one on the same detection pair, within the top 30. Fixes are drawn in their cause's component color, with thick edges and outlined added boxes. Other triplets and objects are grey, and objects the baseline did not box are dashed. R@100 is the per-image \gc R@100. The right column gives the cause, its component and the fix.}
\label{fig:qual}
\end{figure*}
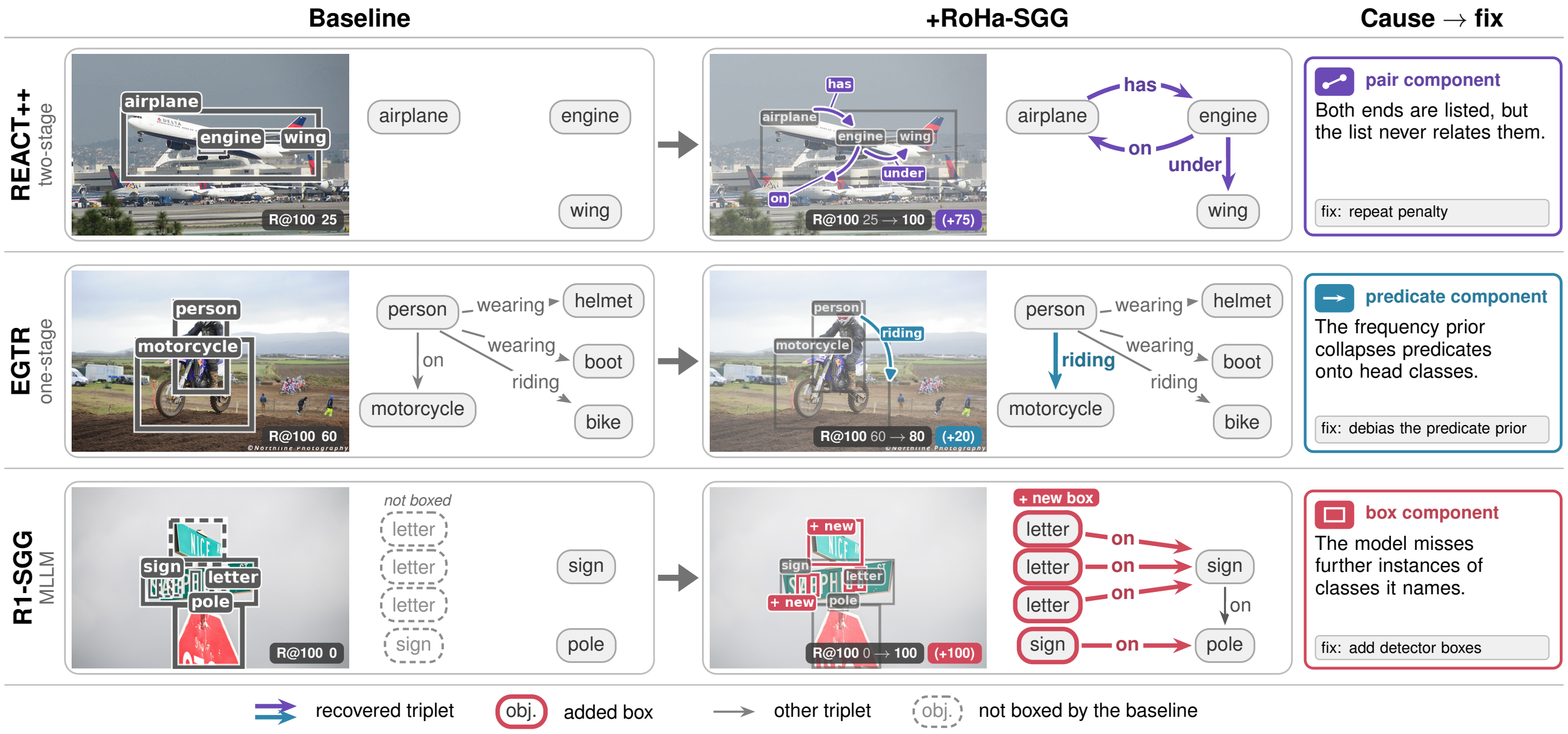

\textbf{Every harness improves every metric.} In all ten searches, it beats its baseline in \cref{tab:main} on R@$K$ and mR@$K$ at $K{=}20$, 50 and 100, F@100 and detection AP50.
For REACT++ and EGTR, single-pass harnesses raise \gc R@100 by 1.6--3.3 points and \gc mR@100 by 0.7--8.7 points, and multi-pass harnesses by 1.2--3.6 and 1.7--10.1 points.
For R1-SGG, single-pass harnesses raise \gc R@100 and \gc mR@100 by up to 10.3 and 20.4 points, and multi-pass harnesses by up to 15.8 and 26.4 points.
On PSG, the multi-pass harness raises R1-SGG's \gc R@100 from 28.4 to 44.2 and \gc mR@100 from 11.5 to 37.9, by correcting predicates and by adding objects that the model leaves without a box (\cref{sec:lossmap}).
Extra model calls mainly help mR@100 (inference time in \cref{tab:time}). Multi-pass harnesses raise \gc mR@100 over single-pass ones in four of the five combinations (by 1.4--6.0 points), and for R1-SGG on VG150 also \gc R@100 by 6.5 points.

Two checks show that the gains come from the harness search itself.
First, the gains do not depend on a single selected harness. Three further harnesses per search, chosen by a pre-specified rule, also raise \gc R@100, mR@100 and F@100 in all 30 cases.
Second, no SGG model is retrained. Most harnesses fit small add-on modules on training images, and most keep their R@100 gain when these modules are removed.

\subsection{Where Models Fail and What Harnesses Fix}
\label{sec:lossmap}
\Cref{fig:lossmap} shows where the baselines fail, as the points of \gc R@20 and mR@20 that each baseline loses at each component.
In \gc R@20, all five baselines lose the most at the pair component. The other components differ by model.
R1-SGG loses far more at the box component because it leaves many objects without a box. REACT++ and EGTR lose more at the rank component, because they output long lists of scored candidate triplets and correct triplets often fall outside the top 20.
To test these differences, we compare the component shares of every two models on the same dataset, three pairs on VG150 and one on PSG, with paired 95\% intervals. For example, on VG150 the box component accounts for 35.6\% of R1-SGG's gain of correcting all components and 9.8\% of EGTR's. The difference is 25.8 percentage points (25.1 to 26.5).
Of the 20 differences per metric, the interval excludes zero for 18 in \gc R@20 and 13 in \gc mR@20, so the models differ significantly at most components.
The box and rank differences hold at $K{=}50$ as well.
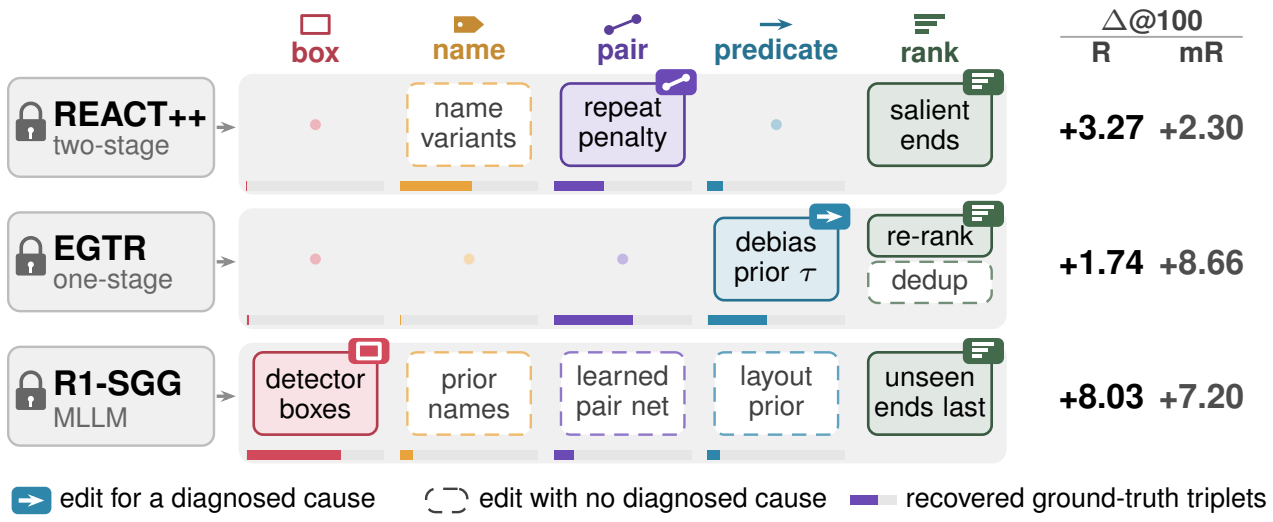
\begin{figure}[t]\setlength{\abovecaptionskip}{0.9em}%
\centering
\colorlet{stRankD}{stRank!55!black}%
\newcommand{\ibadge}[2]{\path ($(c.north east)+(-0.07,0.0)$) coordinate (bc);
  \fill[#2, rounded corners=1.3pt] ($(bc)+(-0.15,-0.095)$) rectangle ($(bc)+(0.15,0.095)$);
  \begin{scope}[shift={(bc)}]\tzStage{#1}{0}{0}{0.72}{white}\end{scope}}%
\resizebox{\columnwidth}{!}{%
\begin{tikzpicture}[font=\sffamily, text=black, line join=round,
  hf/.style={font=\sffamily\fontsize{5.6}{6.2}\selectfont},
  chip/.style={draw=#1!85!black, fill=#1!16, line width=0.55pt, rounded corners=2pt, text width=0.84cm, minimum width=0.90cm, minimum height=0.60cm, inner sep=0.5pt, align=center, hf, text=black},
  hchip/.style={chip=#1, draw=#1!75, fill=white, densely dashed, line width=0.5pt, text=black!75},
  halfchip/.style={chip=#1, minimum height=0.30cm},
  hdr/.style={font=\sffamily\bfseries\fontsize{5.8}{6}\selectfont, text=#1!85!black, inner sep=0pt},
  gh/.style={font=\sffamily\bfseries\fontsize{5.8}{6}\selectfont, text=black!75, inner sep=0pt},
  num/.style={font=\sffamily\bfseries\fontsize{7}{8}\selectfont, text=black}]
\def\XO{1.68}\def\W{1.12}\def\GA{7.97}\def\GB{8.70}
\pgfmathsetmacro{\Xb}{\XO+0.5*\W}\pgfmathsetmacro{\Xn}{\XO+1.5*\W}\pgfmathsetmacro{\Xp}{\XO+2.5*\W}\pgfmathsetmacro{\Xq}{\XO+3.5*\W}\pgfmathsetmacro{\Xr}{\XO+4.5*\W}
\begin{scope}[shift={(\Xb,0.20)}]\draw[stBox!85!black, line width=0.6pt] (-0.10,-0.065) rectangle (0.10,0.065);\end{scope}
\begin{scope}[shift={(\Xn,0.20)}]\fill[stName!85!black] (-0.11,-0.06) -- (0.04,-0.06) -- (0.11,0) -- (0.04,0.06) -- (-0.11,0.06) -- cycle;\fill[white] (-0.065,0) circle (0.018);\end{scope}
\begin{scope}[shift={(\Xp,0.20)}]\draw[stPair!85!black, line width=0.6pt] (-0.10,-0.04) -- (0.10,0.04);\fill[stPair!85!black] (-0.10,-0.04) circle (0.036) (0.10,0.04) circle (0.036);\end{scope}
\begin{scope}[shift={(\Xq,0.20)}]\draw[stPred!85!black, line width=0.7pt, -{Stealth[length=1.1mm]}] (-0.12,0) -- (0.12,0);\end{scope}
\begin{scope}[shift={(\Xr,0.20)}]\foreach \i/\l in {0/0.22,1/0.16,2/0.10} \fill[stRankD] (-0.11,0.045-\i*0.055) rectangle ++(\l,0.032);\end{scope}
\node[hdr=stBox]  at (\Xb,0.0) {box};
\node[hdr=stName] at (\Xn,0.0) {name};
\node[hdr=stPair] at (\Xp,0.0) {pair};
\node[hdr=stPred] at (\Xq,0.0) {predicate};
\node[hdr=stRankD] at (\Xr,0.0) {rank};
\node[gh] at ({(\GA+\GB)/2},0.23) {$\Delta$@100};
\draw[black!35, line width=0.4pt] ({\GA-0.3},0.11) -- ({\GB+0.3},0.11);
\node[gh] at (\GA,0.0) {R};
\node[gh] at (\GB,0.0) {mR};

\newcommand{\hrow}[9]{%
  \node[draw=edgeGrey, fill=panelGrey, rounded corners=2.5pt, line width=0.5pt, minimum width=1.50cm, minimum height=0.72cm] at (0.75,#1) {};
  \begin{scope}[shift={(0.165,#1+0.02)}]%
    \draw[black!60, line width=0.75pt] (-0.07,0.02) -- (-0.07,0.085) arc (180:0:0.07) -- (0.07,0.02);
    \fill[black!60, rounded corners=0.6pt] (-0.11,-0.12) rectangle (0.11,0.04);
    \fill[white] (0,-0.04) circle (0.022); \fill[white] (-0.009,-0.10) rectangle (0.009,-0.045);
  \end{scope}
  \node[anchor=west, inner sep=0pt, font=\sffamily\bfseries\fontsize{7}{8}\selectfont, text=black] at (0.32,#1+0.10) {#2};
  \node[anchor=west, inner sep=0pt, hf, text=black!60] at (0.32,#1-0.15) {#3};
  \draw[black!45, line width=0.7pt, -{Stealth[length=1.1mm]}] (1.52,#1) -- (\XO-0.02,#1);
  \fill[panelGrey, rounded corners=3pt] (\XO,#1-0.49) rectangle (\XO+5*\W,#1+0.39);
  \foreach \s/\c [count=\k from 0] in {#4/stBox,#5/stName,#6/stPair,#7/stPred} {%
    \fill[black!10] (\XO+\k*\W+0.06,#1-0.46) rectangle ++(\W-0.12,0.07);
    \fill[\c] (\XO+\k*\W+0.06,#1-0.46) rectangle ++({(\W-0.12)*\s},0.07);}
  \node[num] at (\GA,#1) {#8};
  \node[num, text=black!70] at (\GB,#1) {#9};
}
\newcommand{\nochip}[2]{\fill[#1!40] (#2) circle (0.04);}

\def\y{-0.55}
\hrow{\y}{REACT++}{two-stage}{0.00}{0.52}{0.36}{0.11}{+3.27}{+2.30}
\nochip{stBox}{\Xb,\y+0.02}
\node[hchip=stName] (c) at (\Xn,\y+0.02) {name\\variants};
\node[chip=stPair] (c) at (\Xp,\y+0.02) {repeat\\penalty};
\ibadge{3}{stPair}
\nochip{stPred}{\Xq,\y+0.02}
\node[chip=stRankD] (c) at (\Xr,\y+0.02) {salient\\ends};
\ibadge{5}{stRankD}
\def\y{-1.53}
\hrow{\y}{EGTR}{one-stage}{0.01}{0.00}{0.57}{0.43}{+1.74}{+8.66}
\nochip{stBox}{\Xb,\y+0.02}
\nochip{stName}{\Xn,\y+0.02}
\nochip{stPair}{\Xp,\y+0.02}
\node[chip=stPred] (c) at (\Xq,\y+0.02) {debias\\prior $\tau$};
\ibadge{4}{stPred}
\node[halfchip=stRankD] (c) at (\Xr,\y+0.02+0.17) {re-rank};
\ibadge{5}{stRankD}
\node[halfchip=stRankD, draw=stRankD!75, fill=white, densely dashed, line width=0.5pt, text=black!75] (c) at (\Xr,\y+0.02-0.17) {dedup};
\def\y{-2.51}
\hrow{\y}{R1-SGG}{MLLM}{0.68}{0.09}{0.14}{0.09}{+8.03}{+7.20}
\node[chip=stBox] (c) at (\Xb,\y+0.02) {detector\\boxes};
\ibadge{1}{stBox}
\node[hchip=stName] (c) at (\Xn,\y+0.02) {prior\\names};
\node[hchip=stPair] (c) at (\Xp,\y+0.02) {learned\\pair net};
\node[hchip=stPred] (c) at (\Xq,\y+0.02) {layout\\prior};
\node[chip=stRankD] (c) at (\Xr,\y+0.02) {unseen\\ends last};
\ibadge{5}{stRankD}
\begin{scope}[shift={(0.17,-3.27)}]\fill[stPred, rounded corners=1.3pt] (-0.145,-0.10) rectangle (0.145,0.10); \tzStage{4}{0}{0}{0.72}{white}\end{scope}
\node[anchor=west, hf, text=black, inner sep=0pt] (l0) at (0.37,-3.27) {edit for a diagnosed cause};
\node[hchip=black, text width=, minimum width=0.30cm, minimum height=0.17cm, right=10pt of l0] (l1) {};
\node[anchor=west, hf, text=black, right=2pt of l1, inner sep=0pt] (l2) {edit with no diagnosed cause};
\fill[black!10] ($(l2.east)+(0.16,-0.035)$) coordinate (l5) rectangle ++(0.34,0.07); \fill[stPair] (l5) rectangle ++(0.2,0.07);
\node[anchor=west, hf, text=black, inner sep=0pt] at ($(l5)+(0.40,0.035)$) {recovered ground-truth triplets};
\end{tikzpicture}}
\caption{\textbf{Selected harnesses for the three models} (VG150, single-pass, \testsplit). Dots mark unchanged components and chips mark edited ones. A badge marks an edit that follows a diagnosed cause. Dashed chips mark edits with no diagnosed cause in the harness's records. They come from free candidates or from earlier steps, which may themselves have followed a cause. Bars show the share of recovered ground-truth triplets that the baseline first lost at each component. The right column shows \testsplit \gc $\Delta$R@100 and $\Delta$mR@100.}
\label{fig:harness}
\end{figure}

The selected harnesses reflect these differences (\cref{fig:harness}).
REACT++'s harnesses recover triplets mainly at the name and pair components, EGTR's at the pair and predicate components, and R1-SGG's at the box component.
\Cref{fig:qual} shows one example per model, from the diagnosed cause to the fix.
For REACT++, the output often lists both objects of a relationship but never relates them (pair component). The selected harness down-weights pairs that are already listed, so that new pairs such as the engine and the wing enter the list.
For EGTR, the frequency prior pushes predicates toward head classes (predicate component). The selected harness debiases this prior and turns a generic ``on'' into the annotated ``riding''.
The two EGTR harnesses raise mR@100 on body and tail predicates by up to 10.8 and 13.2 points, against at most 2.5 on head predicates. The VG150 predicate groups follow BGNN~\cite{li2021bgnn} by training instances (head ${>}10$k, body 0.5--10k, tail ${<}0.5$k).
For R1-SGG, the model misses further instances of the classes it names (box component). The selected harness adds boxes from an external detector and recovers a sign and its letters.

\begin{figure}[t]
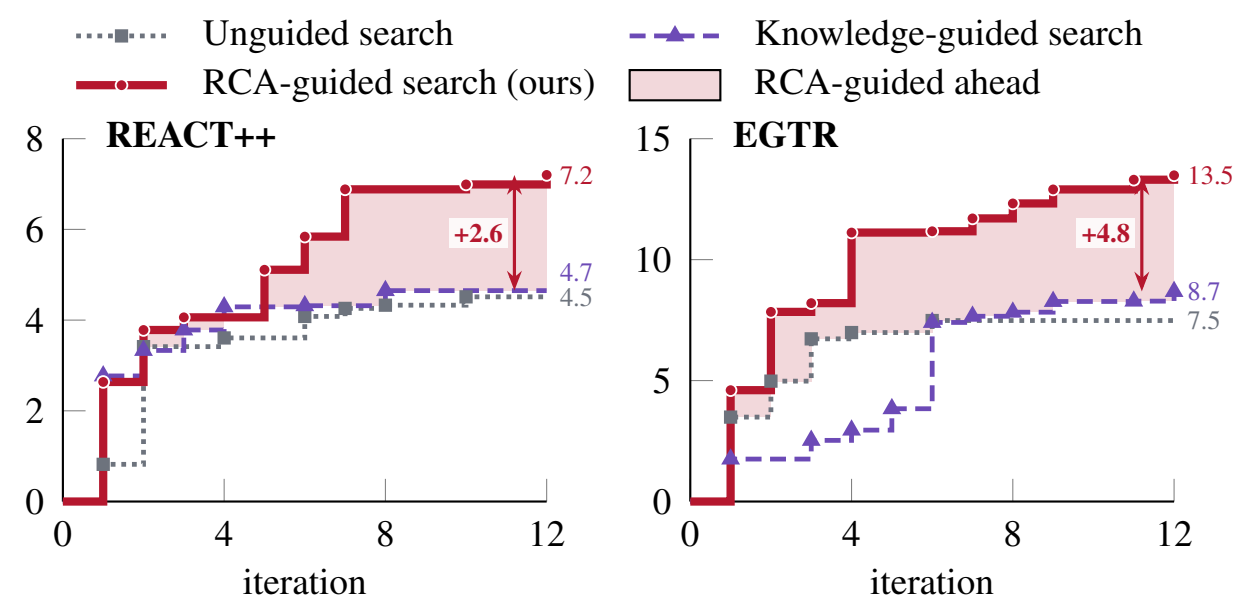

\centering
\colorlet{cbGrey}{search!75}\colorlet{cbBlue}{stPair}\colorlet{cbVerm}{rohared}\colorlet{lvLead}{rohared!17}%
\pgfplotsset{
  lvZ/.style={cbGrey, line width=0.9pt, dash pattern=on 0.9pt off 1.1pt, mark=square*, mark size=0.8pt, mark options={solid}},
  lvO/.style={cbBlue, line width=0.9pt, dash pattern=on 3pt off 1.6pt, mark=triangle*, mark size=1.15pt, mark options={solid}},
  lvT/.style={cbVerm, line width=1.55pt, solid, mark=*, mark size=1.1pt, mark options={fill=cbVerm, draw=white, line width=0.35pt}}}
\tikzset{lvZLab/.style={text=cbGrey}, lvOLab/.style={text=cbBlue}, lvTLab/.style={text=cbVerm, font=\tiny\bfseries}}
\pgfplotsset{every tick label/.append style={font=\rmfamily}, /pgf/number format/assume math mode=true}%
\begin{tikzpicture}
\begin{groupplot}[group style={group size=2 by 1, horizontal sep=0.95cm},
  scale only axis, width=3.2cm, height=2.4cm, clip=false, xmin=0, xmax=36, ymin=0, xtick={0,12,24,36}, xticklabels={0,4,8,12},
  x tick label style={font=\scriptsize}, y tick label style={font=\scriptsize}, title style={font=\scriptsize\bfseries, at={(rel axis cs:0.07,1)}, anchor=north west, inner sep=1.5pt, yshift=-1pt},
  xlabel={\scriptsize iteration}, xlabel style={yshift=0.6ex},
  axis x line*=bottom, axis y line*=left]
\nextgroupplot[title={REACT++}, ymax=8, ytick={0,2,4,6,8}]
\input{figures/generated/levels_v3_k100_React}
\nextgroupplot[title={EGTR}, ymax=15, ytick={0,5,10,15},
  legend style={font=\scriptsize, at={(-0.16,1.04)}, anchor=south, legend columns=2, draw=none, column sep=4pt, row sep=-1pt, fill=none}, legend cell align=left]
\addlegendimage{lvZ}\addlegendentry{Unguided search}
\addlegendimage{lvO}\addlegendentry{Knowledge-guided search}
\addlegendimage{lvT}\addlegendentry{RCA-guided search (ours)}
\addlegendimage{area legend, fill=lvLead, draw=none}\addlegendentry{RCA-guided ahead}
\input{figures/generated/levels_v3_k100_EGTR}
\end{groupplot}
\end{tikzpicture}
\caption{\textbf{Unguided, knowledge-guided and RCA-guided search} (VG150, single-pass). Best harness so far (by the ranking of \cref{sec:search}) per iteration, as a running maximum. The $y$-axis shows the gain over the baseline, $\Delta$R@100$+\Delta$mR@100 (\gc, \searchsplit split). The shaded area marks the lead of the RCA-guided search over the better other setting, and the arrow marks its final lead.}
\label{fig:levels}
\end{figure}

\subsection{Effect of Root-Cause Analysis}
\label{sec:levels}
To measure the effect of root-cause analysis, we compare three search settings on the VG150 single-pass searches of all three models.
All three settings run 12 iterations with three candidate harnesses each, and they share the model interface, tools, scoring and selection rule.
The \emph{unguided search} uses only the search loop adapted from Meta-Harness~\cite{lee2026metaharness}, so its proposer receives no diagnosed causes.
The \emph{knowledge-guided search} also gives the proposer a human-written guide based on the SGG literature.
The \emph{RCA-guided search} instead gives the proposer the causes diagnosed by the RCA agent, as described in \cref{sec:method}.

\paragraph{Search efficiency.}
On REACT++ and EGTR, the RCA-guided search reaches the final score of the unguided and knowledge-guided searches after only 5 and 4 of 12 iterations, up to three times faster (\cref{fig:levels}).
For EGTR, it leads at every iteration, almost entirely through \gc mR@100.
For REACT++, the gap opens with the first re-diagnosis. During the second round, the new causes lift the score from 4.1 to 6.9, while the other two settings stay nearly flat.
Candidates guided by the diagnosis also reduce the error they target more often. In the first round, 60\% of them lower the error of their target component, against 41\% when targets are randomly permuted within a search (permutation test, $p<0.001$).

\paragraph{Test results.}
At iteration 12, the RCA-guided search also gives the largest gain in $\mu$ for EGTR (+10.4 vs.\ +7.8 and +7.5), mostly from \gc mR@100 and with the highest F@100, and for REACT++ (+5.6 vs.\ +3.7 and +3.3).

\begin{table}[t]\centering\footnotesize\setlength{\tabcolsep}{3pt}
\caption{\textbf{Search cost.} API spend of one RCA-guided search in US dollars, including its root-cause analyses (\$62--105). The harness makes no API call at inference.}\label{tab:searchcost}
\begin{tabular*}{\columnwidth}{@{\extracolsep{\fill}}lccccc@{}}\toprule
 & \multicolumn{3}{c}{VG150} & \multicolumn{2}{c}{PSG}\\\cmidrule(lr){2-4}\cmidrule(l){5-6}
Search & REACT++ & EGTR & R1-SGG & REACT++ & R1-SGG\\\midrule
Single-pass & \$270 & \$325 & \$347 & \$263 & \$320\\
Multi-pass & \$318 & \$314 & \$376 & \$295 & \$296\\
\bottomrule\end{tabular*}\end{table}

\subsection{Discussion and Limitations}
\label{sec:limits}
\paragraph{When does the analysis help?}
The analysis helps most when a model's errors cannot be seen from its output or its recall alone.
For R1-SGG, the three settings end within 1.7 points on the search split.
R1-SGG leaves many objects without a box, an error that is visible in its output. Every setting therefore adds detector boxes, and the analysis adds little beyond that.
REACT++, in contrast, also loses triplets at object names, an error that its recall alone does not show.
Only the RCA-guided search recovers these name errors (52\% of its recovered triplets), through a candidate that read the errors per component, and it raises \gc mR@100 two to five times as much as the other searches.

\paragraph{Inference and search cost.}
Single-pass harnesses add under 1\,ms per image for EGTR, about 13\% for R1-SGG and under 1\,s for REACT++, which calls external detectors (\cref{tab:time}).
A search costs \$263--376 in API fees, including \$62--105 for the analysis (\cref{tab:searchcost}). The analysis pays for itself. On REACT++ and EGTR, the RCA-guided search reaches the final score of the other two searches early. Up to that point, it spends less than those searches spend in total, including all its analysis costs.

\paragraph{Limitations.}
Each search runs once, so the intervals reflect image, not run-to-run, variation~\cite{bouthillier2021variance}. Other LLM agents and the PredCls and SGCls settings are future work.

\section{Conclusion}
\label{sec:conclusion}
We presented \methodname, the first agentic harness optimization for SGG.
Its RCA agent finds where and why a frozen SGG model fails, and its proposer agent automatically searches for harnesses that fix them without retraining.
The analysis shows that models lose ground-truth triplets at different components.
The searched harnesses improve every metric on all five model--dataset combinations, across two-stage, one-stage and multimodal-LLM models. The gains reach 15.8 \gc R@100 and 26.4 \gc mR@100 points.
Root-cause analysis also makes the search up to three times faster than unguided and knowledge-guided searches.

\FloatBarrier
{
    \small
    \bibliographystyle{ieeenat_fullname}
    \bibliography{main}
}

\end{document}